%% file: top.tex
\documentclass[10pt,twocolumn,letterpaper]{article}

\usepackage{iccv}
\usepackage{times}
\usepackage{epsfig}
\usepackage{graphicx}
\usepackage{amsmath}
\usepackage{amssymb}

\usepackage[sort,nocompress]{cite}
\usepackage[utf8]{inputenc}
\usepackage{float}
\usepackage{booktabs}
\usepackage{caption}
\usepackage{subcaption}
\usepackage{multirow}

\usepackage{xcolor}
\usepackage{soul}
\usepackage{pifont}
\usepackage[accsupp]{axessibility}
\newcommand{\cmark}{\ding{51}}
\newcommand{\xmark}{\ding{55}}
\usepackage[pagebackref=true,breaklinks=true,letterpaper=true,colorlinks,bookmarks=false]{hyperref}

\iccvfinalcopy %

\def\iccvPaperID{1194} %

\ificcvfinal\fi

\input{shortcuts}

\graphicspath{./gfx/}

\begin{document}

\title{UNISURF: Unifying Neural Implicit Surfaces and\\Radiance Fields for Multi-View Reconstruction}

\author{Michael Oechsle$^{1,2,3}$ \quad Songyou Peng$^{1,4}$ \quad Andreas Geiger$^{1,2}$\\
	$^1$Max Planck Institute for Intelligent Systems, Tübingen \qquad $^2$University of Tübingen\\
	$^3$ETAS GmbH, Stuttgart \qquad $^4$ETH Zurich\\
	{\tt\small \{firstname.lastname\}@tue.mpg.de}
	\vspace{-0.3cm}
}

\maketitle

\ificcvfinal\thispagestyle{empty}\fi

\input{sec_abstract}

\input{sec_intro}

\input{sec_related}

\input{sec_method_andreas}
\input{sec_results}

\input{sec_conclusion}

{\small
	\bibliographystyle{ieee_fullname}
	\bibliography{bibliography_long,bibliography,bibliography_custom}
}

\end{document}

%% file: shortcuts.tex
\newcommand{\bd}{\mathbf{d}}

\newcommand{\bh}{\mathbf{h}}

\newcommand{\bn}{\mathbf{n}}
\newcommand{\bo}{\mathbf{o}}

\newcommand{\br}{\mathbf{r}}

\newcommand{\bx}{\mathbf{x}}

\newcommand{\bepsilon}{\boldsymbol{\epsilon}}

\newcommand{\nR}{\mathbb{R}}

\newcommand{\cL}{\mathcal{L}}

\newcommand{\cR}{\mathcal{R}}
\newcommand{\cS}{\mathcal{S}}

\newcommand{\figref}[1]{Fig.~\ref{#1}}
\newcommand{\secref}[1]{Section~\ref{#1}}

\newcommand{\eqnref}[1]{Eq.~\eqref{#1}}
\newcommand{\tabref}[1]{Table~\ref{#1}}

\makeatletter
\DeclareRobustCommand\onedot{\futurelet\@let@token\@onedot}
\def\@onedot{\ifx\@let@token.\else.\null\fi\xspace}
\def\eg{e.g\onedot} 
\def\ie{i.e\onedot}

\def\vs{vs\onedot}
\def\wrt{wrt\onedot}

\def\etal{et~al\onedot}

\makeatother

\newcommand{\boldparagraph}[1]{\vspace{0.2cm}\noindent{\bf #1:}}

\definecolor{darkgreen}{rgb}{0,0.7,0}

%% file: sec_abstract.tex
\begin{abstract}
Neural implicit 3D representations have emerged as a powerful paradigm for reconstructing surfaces from multi-view images and synthesizing novel views. Unfortunately, existing methods such as DVR or IDR require accurate per-pixel object masks as supervision. At the same time, neural radiance fields have revolutionized novel view synthesis. However, NeRF's estimated volume density does not admit accurate surface reconstruction. Our key insight is that implicit surface models and radiance fields can be formulated in a unified way, enabling both surface and volume rendering using the same model. This unified perspective enables novel, more efficient sampling procedures and the ability to reconstruct accurate surfaces without input masks. We compare our method on the DTU, BlendedMVS, and a synthetic indoor dataset. Our experiments demonstrate that we outperform NeRF in terms of reconstruction quality while performing on par with IDR without requiring masks.
\end{abstract}

%% file: sec_intro.tex
\section{Introduction}
\label{sec:intro}
\begin{figure}
    \centering
    \includegraphics[width=\linewidth]{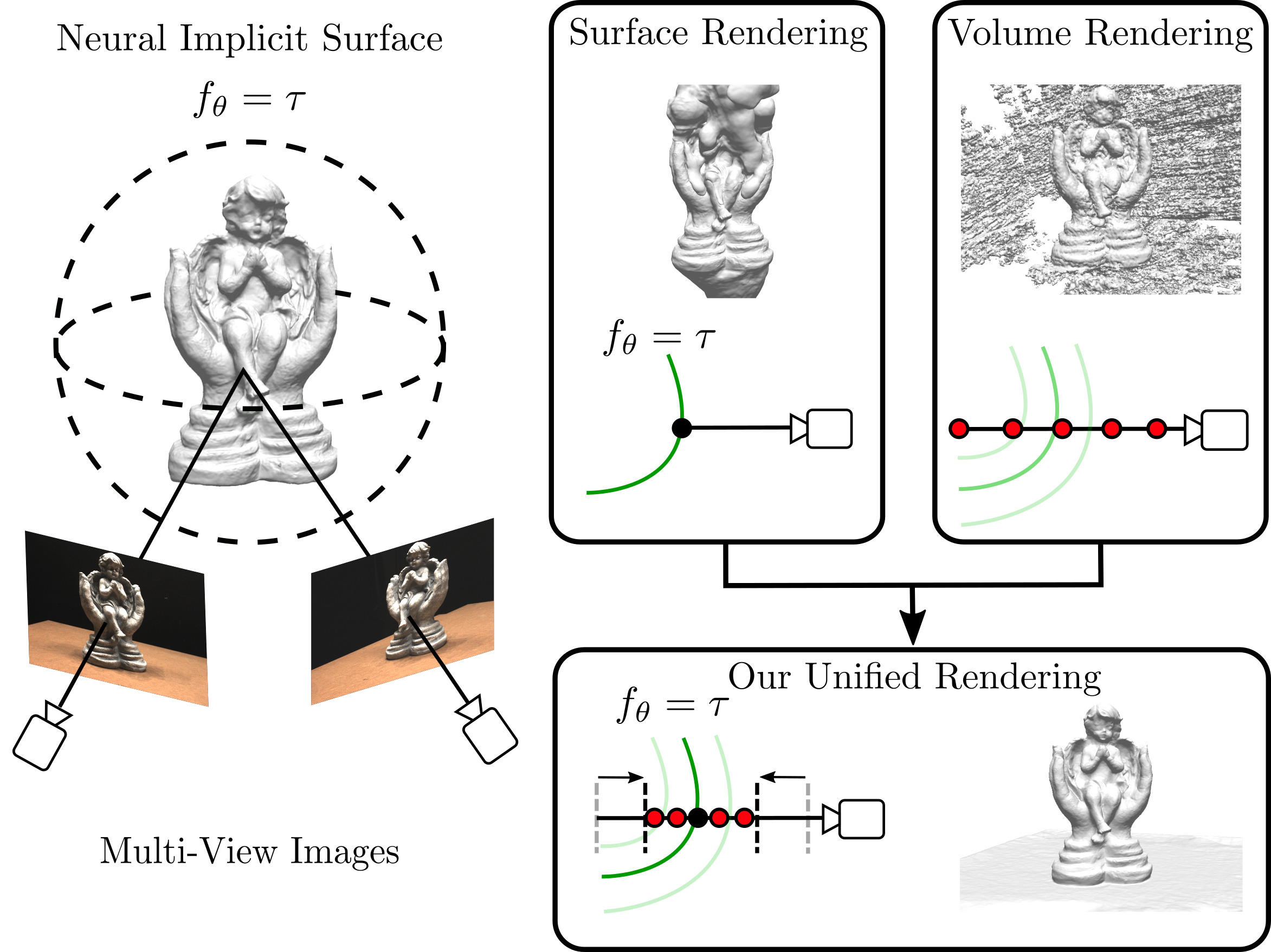}
    \caption{
    \textbf{Illustration.}
    Implicit models based on surface rendering \cite{Yariv2020ARXIV,Niemeyer2020CVPR} require input masks  and radiance fields \cite{Mildenhall2020ECCV} do not optimize implicit surfaces directly. UNISURF provides a principled unified formulation, enabling accurate surface reconstruction from images without input masks.
	}
   \vspace*{-0.3cm}
\end{figure}

Capturing the geometry and appearance of 3D scenes from a set of images is one of the cornerstone problems in computer vision. Towards this goal, coordinate-based neural models have emerged as a powerful tool for 3D reconstruction of geometry and appearance within the last years. %

Many recent methods employ continuous implicit functions parameterized with neural networks as 3D representations of geometry \cite{Mescheder2019CVPR, Chen2018CVPR, Park2019CVPR,Wang2019NIPS, Saito2019ICCV,Michalkiewicz2019ICCV,Genova2019ICCV,Niemeyer2019ICCV,Atzmon2019NIPS,Peng2020ECCV} or appearance \cite{Saito2019ICCV, Niemeyer2020CVPR,Oechsle2019ICCV,Sitzmann2019NIPS,Oechsle2020THREEDV,Mildenhall2020ECCV,Yariv2020ARXIV}. 
These neural 3D representations have shown impressive performance on geometry reconstruction and novel view synthesis from multi-view images. 
Besides the choice of the 3D representation (\eg, occupancy field, unsigned or signed distance field), one key element for neural implicit multi-view reconstruction is the rendering technique.
While some of these works represent the implicit surface as level set and hence render the appearance from surfaces \cite{Niemeyer2020CVPR, Sitzmann2019NIPS, Yariv2020ARXIV}, others integrate densities by drawing samples along the viewing rays \cite{Mildenhall2020ECCV,Liu2020NEURIPS,Schwarz2020NEURIPS}.

\begin{figure}
    \center
    \begin{tabular}{c@{}c@{}c}
       \includegraphics[trim={300 50 300 0},clip,width=0.33\linewidth]{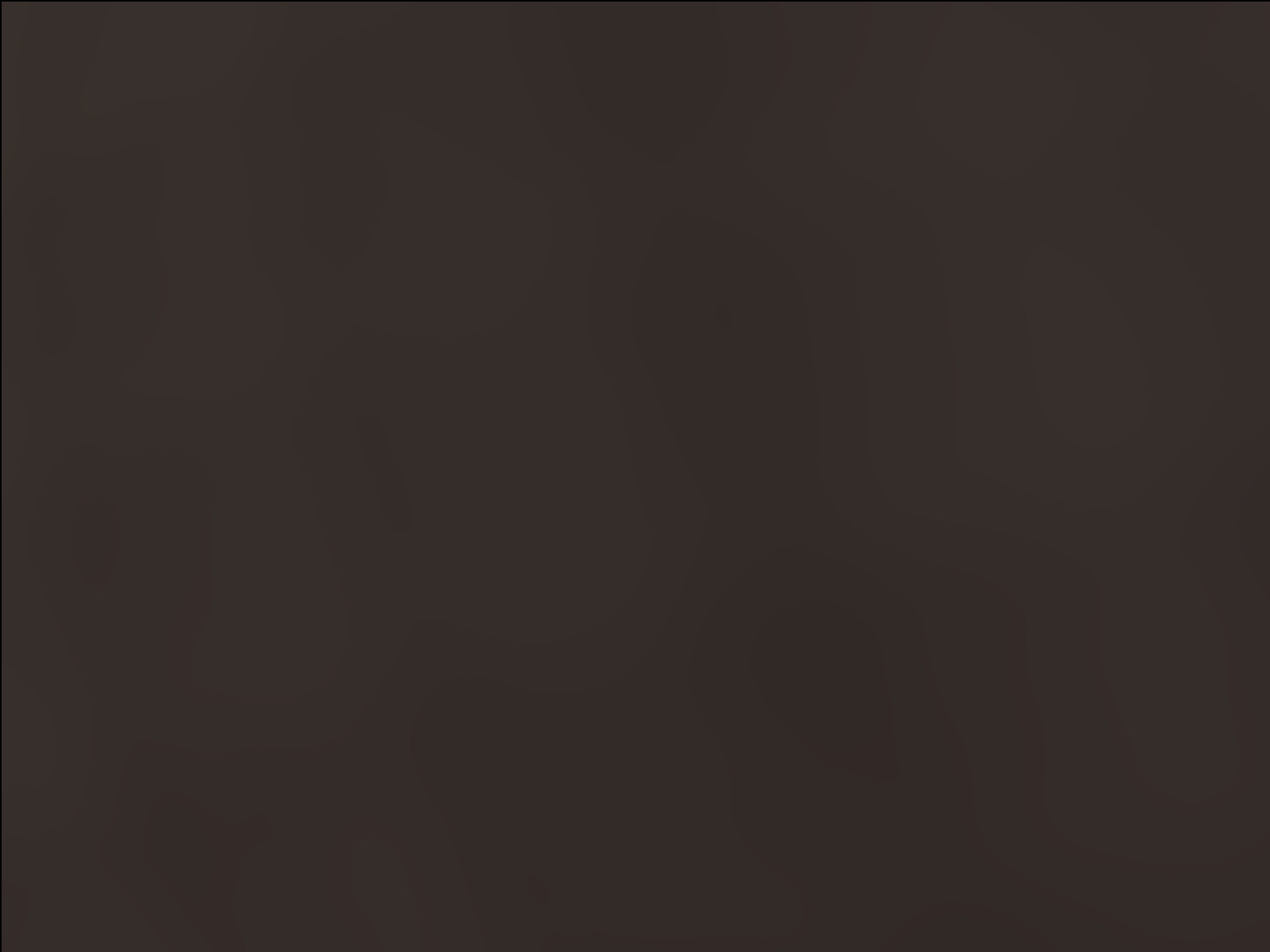}&
       \includegraphics[trim={300 50 300 0},clip,width=0.33\linewidth]{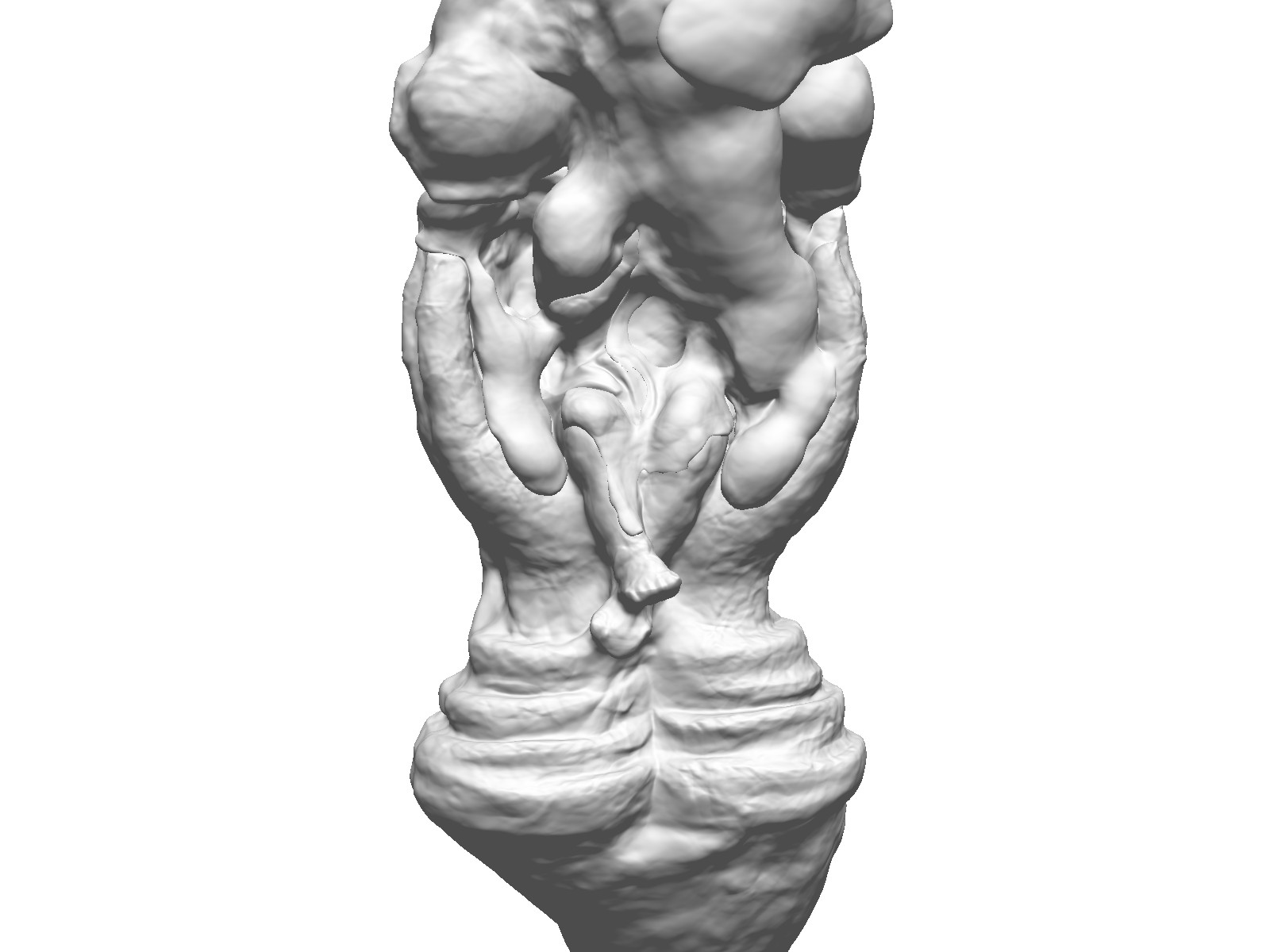}&
       \includegraphics[trim={150 25 150 0},clip,width=0.33\linewidth]{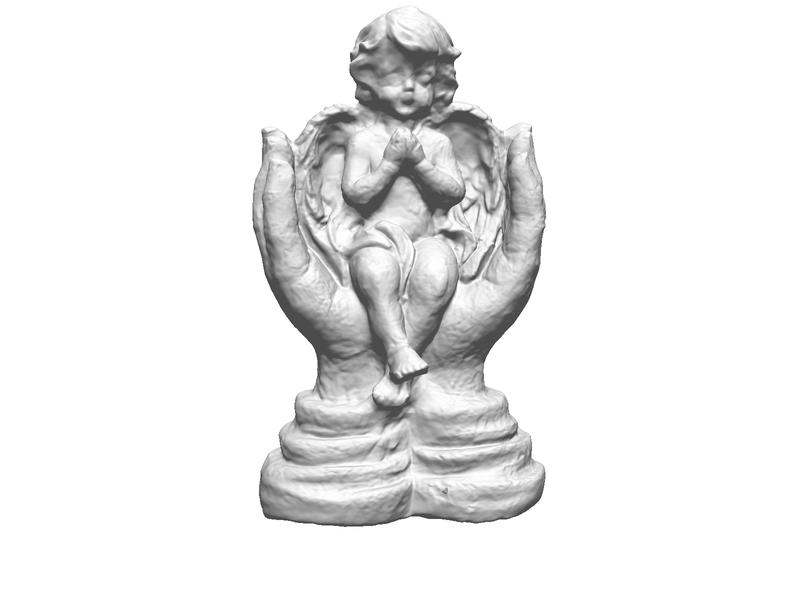}\\
       Random Init. &
       Sphere Init. &
       Sphere Init. \\
      No Masks &
	  No Masks\footnotemark&
   	  With Masks
    \end{tabular}
    \caption{\textbf{Surface Rendering (IDR Results).} State-of-the-art methods like IDR \cite{Yariv2020ARXIV} require object masks and careful initialization for capturing accurate geometry.}
    \label{fig:idr_bounds}
 \end{figure}

In existing work, surface rendering techniques have shown impressive performance in 3D reconstruction \cite{Niemeyer2020CVPR, Yariv2020ARXIV}. 
However, they require per-pixel object masks as input and an appropriate network initialization since surface rendering techniques only provide gradient information locally where a surface intersects with a ray.
Intuitively speaking, optimizing \wrt local gradients can be seen as an iterative deformation procedure applied to an initial neural surface which is often initialized as a sphere. Additional constraints such as mask supervision are necessary for converging to a valid surface, see \figref{fig:idr_bounds} for an illustration.
Due to their reliance on masks, surface rendering methods are limited to object-level reconstruction and do not scale to larger scenes.

\begin{figure}
    \center
    \begin{tabular}{c@{}c@{}c}
       \includegraphics[width=0.32\linewidth]{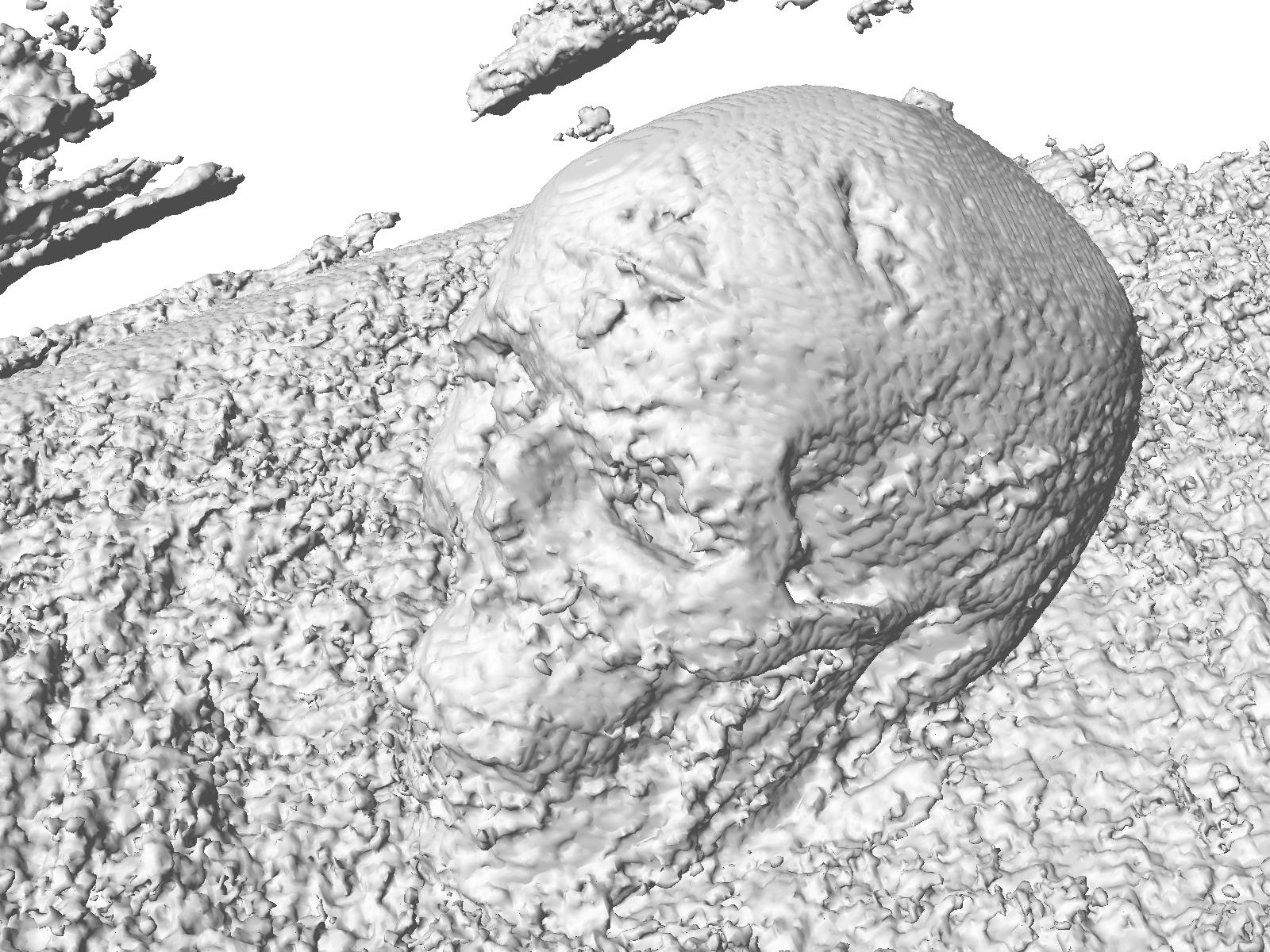}&
       \includegraphics[width=0.32\linewidth]{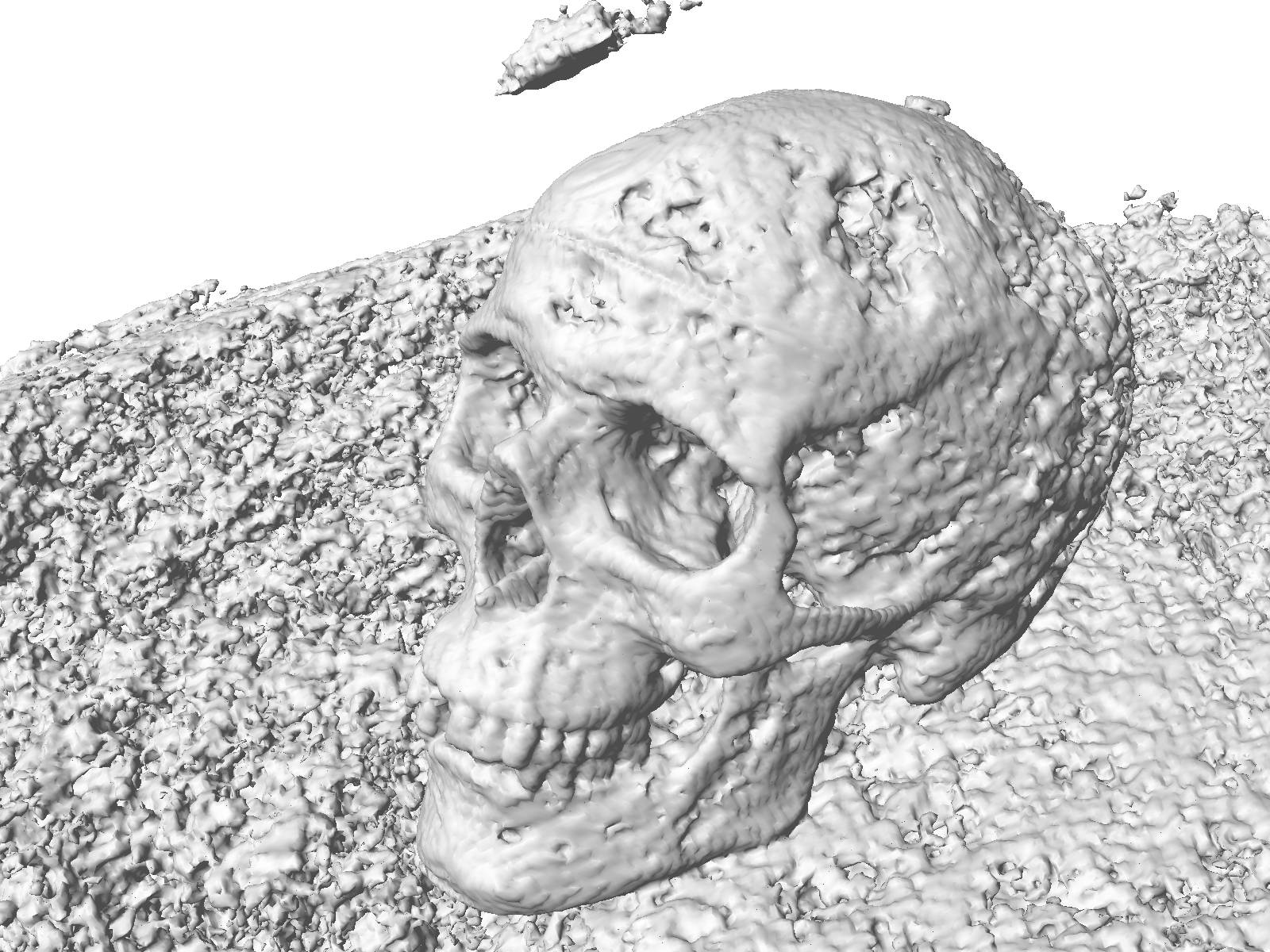}&
       \includegraphics[width=0.32\linewidth]{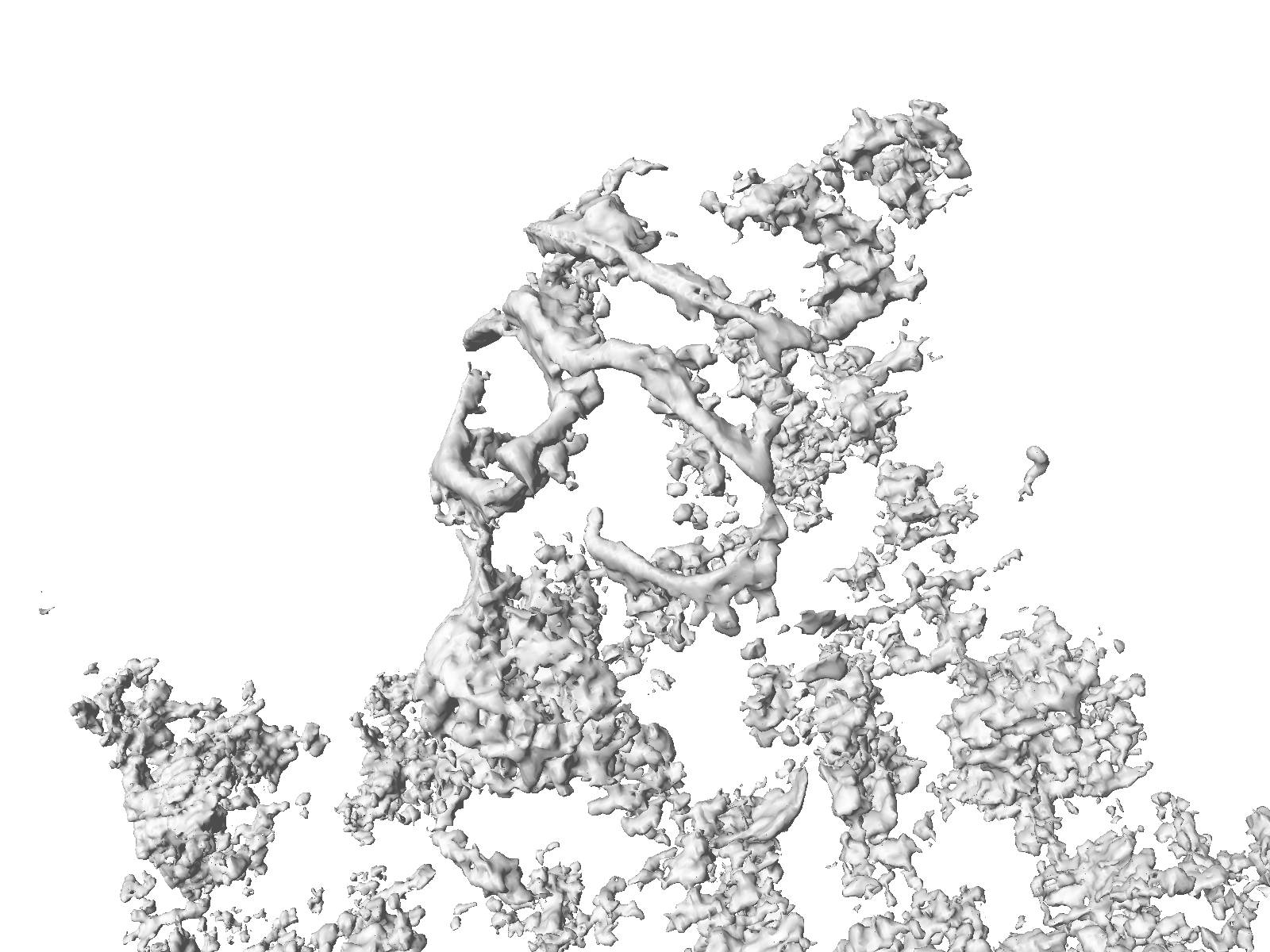}\\
       $\sigma=1$&
       $\sigma=50$&
       $\sigma=500$\\
    \end{tabular}
    \caption{\textbf{Volume Rendering (NeRF Results).} We show level sets of the recovered density volume for a trained NeRF model \cite{Mildenhall2020ECCV} using different density thresholds $\sigma$.}
    \label{fig:nerf_bounds}
    \vspace{-0.2cm}
 \end{figure}

On contrary, volume rendering methods like NeRF \cite{Mildenhall2020ECCV} have shown impressive results for novel view synthesis, also for larger scenes. 
However, surfaces extracted as level sets of the underlying volume density are usually non-smooth and contain artifacts due to the flexibility of the radiance field representation which does not sufficiently constrain the 3D geometry in the presence of ambiguities, see \figref{fig:nerf_bounds}.

 \footnotetext{We remark that this result w/o masks from IDR~\cite{Yariv2020ARXIV} was the only successful case out of several failure cases.}
 
\boldparagraph{Contributions}
In this paper, we propose \textit{UNISURF (UNIfied Neural Implicit SUrface and Radiance Fields)} a principled unified framework for implicit surfaces and radiance fields, with the goal of reconstructing solid (\ie, non-transparent) objects from a set of RGB images.
Our framework combines
the benefits of surface rendering with those of volume rendering, enabling the reconstruction of accurate geometry from 
multi-view images without masks. 
By recovering implicit surfaces, we are able to gradually decrease the sampling region for volume rendering during optimization.
Starting with large sampling regions enables capturing coarse geometry and resolving ambiguities during early iterations. At a later stage, we draw samples closer to the surface which improves reconstruction accuracy.
We show that our approach enables capturing accurate geometry without mask supervision
on the \emph{DTU MVS} dataset~\cite{Aanes2016IJCV},
attaining results competitive with state-of-the-art implicit neural reconstruction methods like IDR \cite{Yariv2020ARXIV} which use strong mask supervision.
Moreover, we also demonstrate our method on scenes from the \emph{BlendedMVS} dataset~\cite{Yao2020CVPR} as well as synthetic indoor scenes from SceneNet~\cite{McCormac2017ICCV}.
Code is available at \url{https://github.com/autonomousvision/unisurf}.

%% file: sec_related.tex
\section{Related Work}
\label{sec:related}
In this section, we first discuss related work from the domain of 3D reconstruction from multi-view images. Next, we provide an overview of recent works on neural implicit representations as well as differentiable rendering.

\boldparagraph{3D Reconstruction from Multi-View Images}
Reconstructing 3D geometry from multiple images has been a longstanding computer vision problem~\cite{Hartley2003}. 
Before the era of deep learning, classic multi-view stereo (MVS) methods~\cite{Agrawal2001CVPR,Bleyer2011BMVC,Bonet1999ICCV,Kutulakos2000IJCV,Broadhurst2001ICCV,Kutulakos2000IJCV,Schoenberger2016ECCV,Seitz1997CVPR,Seitz2006CVPR} focus on either matching features across views~\cite{Bleyer2011BMVC,Schoenberger2016ECCV} or representing shapes with a voxel grid~\cite{Marr1976Science,Agrawal2001CVPR,Bonet1999ICCV,Broadhurst2001ICCV,Seitz1997CVPR,Kutulakos2000IJCV,Paschalidou2018CVPR,Tulsiani2017CVPR,Ulusoy2015THREEDV}. The former approaches usually have a complex pipeline requiring additional steps like fusing depth information~\cite{Curless1996SIGGRAPH,Merrell2007ICCV} and meshing~\cite{Kazhdan2006PSR,Kazhdan2013SIGGRAPH}, while the latter ones are limited to low resolution due to cubic memory requirements.
In contrast, neural implicit representations for 3D reconstruction do not suffer from discretization artifacts as they represent surfaces by the level set of a neural network with continuous outputs.

Recent learning-based MVS methods attempt to replace some parts of the classic MVS pipeline.
For instance, some works learn to match 2D features\cite{Hartmann2017ICCV,Leroy2018ECCV,Luo2016CVPR,Ummenhofer2017CVPRa,Zagoruyko2015CVPR}, fuse depth maps~\cite{Donne2019CVPR,Riegler2017THREEDV}, or infer depth maps~\cite{Huang2018CVPRa,Yao2018ECCV,Yao2019CVPR} from multi-view images.
Contrary to these learning-based MVS approaches, our method only requires weak 2D supervision during optimization. 
Moreover, our method yields high-quality 3D geometry \textit{and} synthesizes photorealistic and consistent novel views.

\boldparagraph{Neural Implicit Representations}
Recently, neural implicit functions have emerged as an effective representation of 3D geometry \cite{Mescheder2019CVPR, Chen2018CVPR, Park2019CVPR,Wang2019NIPS, Saito2019ICCV,Michalkiewicz2019ICCV,Genova2019ICCV,Niemeyer2019ICCV, Atzmon2019NIPS,Peng2020ECCV} and appearance \cite{Saito2019ICCV,Niemeyer2020CVPR,Oechsle2019ICCV,Liu2020CVPR,Sitzmann2019NIPS,Oechsle2020THREEDV,Mildenhall2020ECCV,Liu2020NEURIPS,Schwarz2020NEURIPS} as they represent 3D content continuously and without discretization while simultaneously having a small memory footprint.
Most of these methods require 3D supervision.
However, several recent works~\cite{Niemeyer2020CVPR, Sitzmann2019NIPS,Yariv2020ARXIV,Mildenhall2020ECCV,Liu2019NIPSb} demonstrated differentiable rendering for training directly from images~\cite{Niemeyer2020CVPR, Sitzmann2019NIPS,Yariv2020ARXIV,Mildenhall2020ECCV,Liu2019NIPSb}.
We divide these methods into two groups: surface rendering and volume rendering.

Surface rendering approaches, including DVR~\cite{Niemeyer2020CVPR} and IDR~\cite{Yariv2020ARXIV}, determine the radiance directly on the surface of an object and provide a differentiable rendering formulation using implicit gradients. 
This allows for optimizing neural implicit surfaces from multi-view images.
Conditioning on the viewing direction allows IDR to capture a high level of detail, even in the presence of non-lambertian surfaces.
However, both DVR and IDR require pixel-accurate object masks for all views as input.
In contrast, our method leads to similar reconstructions without requiring masks.

NeRF~\cite{Mildenhall2020ECCV} and follow-ups \cite{Zhang2020nerfpp,Srinivasan2021CVPR,Martinbrualla2021CVPR,Schwarz2020NEURIPS,Niemeyer2021CVPR,Neff2021donerf,Peng2021CVPR,Boss2020ARXIV,Pumarola2020ARXIV} use volume rendering by learning alpha-compositing of a radiance field along rays. 
This method has shown impressive results on novel view synthesis and does not require mask supervision.
However, the recovered 3D geometry is far from satisfactory, see \figref{fig:nerf_bounds}.
Several follow-up works (Neural Body~\cite{Peng2021CVPR} D-NeRF~\cite{Pumarola2020ARXIV} and NeRD~\cite{Boss2020ARXIV}) extract meshes using the volume density from NeRF, but none of them considers optimizing surfaces directly.
Unlike these works, we aim at capturing accurate geometry and propose a volume rendering formulation that provably approaches surface rendering in the limit.

%% file: sec_method_andreas.tex
\section{Background}
\label{sec:background}

The two main ingredients for learning neural implicit 3D representations from multi-view images are the 3D representation and the rendering technique linking the 3D representation and the 2D observations.
This section provides the relevant background on implicit surface and volumetric radiance representations which we unify in this paper for the case of solid (non-transparent) objects and scenes.

\boldparagraph{Implicit Surface Models}
Occupancy Networks \cite{Mescheder2019CVPR,Niemeyer2020CVPR} represent surfaces as the decision boundary of a binary occupancy classifier, parameterized by a neural network
\begin{equation}
o_\theta(\bx): \mathbb{R}^3 \rightarrow [0,1]
\label{eq:onet}
\end{equation}
where $\bx\in\nR^3$ is a 3D point and $\theta$ are the model parameters. The surface is defined as the set of all 3D points where the occupancy probability is one half: $\cS=\{\bx_s|o_\theta (\bx_s) = 0.5\}$.
To associate a color with every 3D point $\bx_s$ on the surface, a color field $c_\theta(\bx_s)$ can be learned jointly with the occupancy field $o_\theta(\bx)$. The color for a particular pixel/ray $\br$ is thus predicted as follows
\begin{align}
\hat{C}(\br) &= c_\theta (\bx_s)
\end{align}
where $\bx_s$ is retrieved by root finding along ray $\br$, see \cite{Niemeyer2020CVPR} for details.
The parameters\footnote{For convenience, we use the same symbol $\theta$ for all model parameters.} $\theta$ of the occupancy field $o_\theta$ and the color field $c_\theta$ are determined by optimizing a reconstruction loss via gradient descent as described in \cite{Niemeyer2020CVPR,Liu2020CVPR,Yariv2020ARXIV}.
 
While surface rendering allows for accurately estimating geometry and appearance, existing approaches strongly rely on supervision with object masks as surface rendering methods are only able to reason about rays that intersect a surface.

\boldparagraph{Volumetric Radiance Models}\label{sec:volumerendering}
In contrast to implicit surface models, NeRF~\cite{Mildenhall2020ECCV} represents scenes as colorized volume densities and integrates radiance along rays via alpha blending~\cite{Lombardi2019SIGGRAPH,Mildenhall2020ECCV}.
More specifically, NeRF uses a neural network to map a 3D location $\bx\in\nR^3$ and a viewing direction $\bd\in\nR^3$ to a volume density $\sigma_\theta (\bx)\in\nR^+$ and a color value $c_\theta(\bx, \bd)\in\nR^3$.
Conditioning on the viewing direction $\bd$ allows for modeling view-dependent effects such as specular reflections \cite{Oechsle2020THREEDV,Mildenhall2020ECCV} and improves reconstruction quality in case the Lambertian assumption is violated \cite{Yariv2020ARXIV}.
Let $\bo$ denote the location of the camera center.
Given $N$ samples $\{\bx_i\}$ along ray $\br=\bo + t \bd$, NeRF approximates the color of pixel/ray $\br$ using numerical quadrature:
\begin{align}
\hat{C}(\br) &= \sum_{i=1}^{N} T_i \, \left(1 - \exp\left( - \sigma_\theta(\bx_i)\,\delta_i \right)\right) \, c_\theta(\bx_i, \bd)
\label{eq:nerf_original}\\
T_i &= \exp \left(- \sum_{j<i} \sigma_\theta(\bx_j) \, \delta_j\right)
\label{eq:nerf_transmittance}
\end{align}
Here, $T_i$ is the accumulated transmittance along the ray and $\delta_i = {\vert \bx_{i+1} -\bx_{i}\vert}$ is the distance between adjacent samples.
As \eqnref{eq:nerf_original} is differentiable, the parameters $\theta$ of the density field $\sigma_\theta$ and the color field $c_\theta$ can be estimated by optimizing a reconstruction loss. We refer to \cite{Mildenhall2020ECCV} for details.

While NeRF does not require object masks for training due to its volumetric radiance representation, extracting the scene geometry from the volume density requires careful tuning of the density threshold and leads to artifacts due to the ambiguity present in the density field, see \figref{fig:nerf_bounds}.

\section{Method}
\label{sec:method}
\begin{figure}
    \centering
    \includegraphics[width=0.8\linewidth]{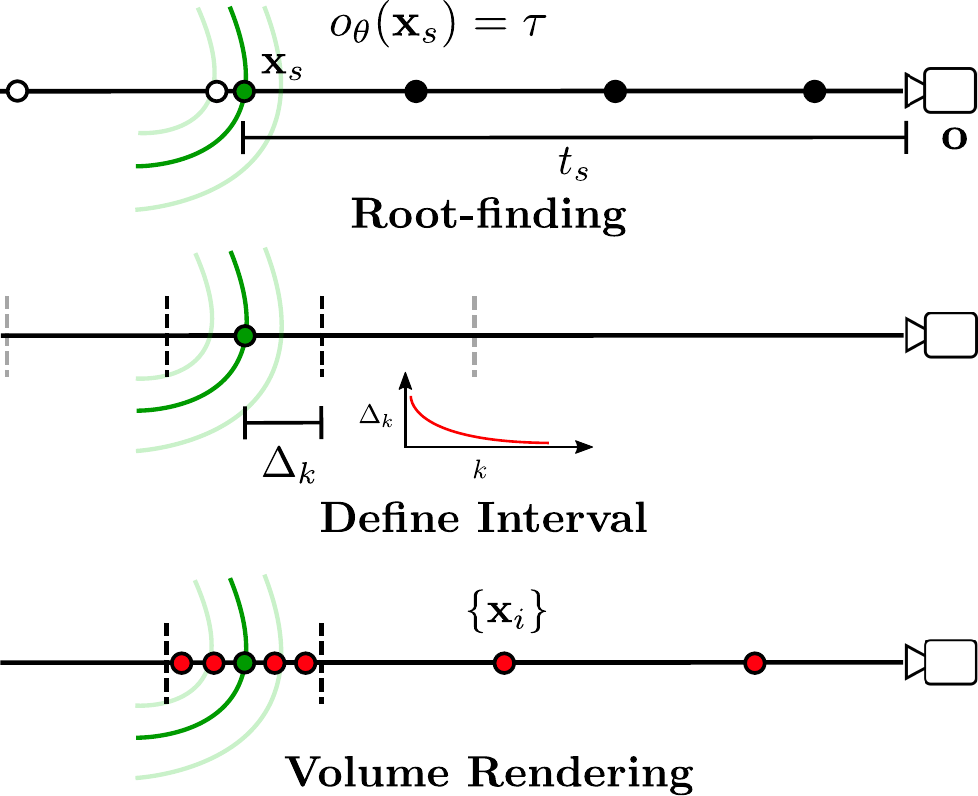}
    \caption{\textbf{Concept and Notation.} Our rendering consists of two steps: First, we seek the surface $\bx_s$ (\textcolor{darkgreen}{green}) in the occupancy field $o_\theta$. 
	Second, we define an interval around the surface to sample points $\{\bx_i\}$ (\textcolor{red}{red}) for volume rendering.
	}
	\label{fig:concept}
\end{figure}

We now describe our main contribution.
In contrast to NeRF which is also applicable to non-solid scenes (\eg, fog, smoke), we restrict our focus to solid objects that can be represented by 3D surfaces and view-dependent surface colors. Our method exploits both, the power of volumetric radiance representations to learn coarse scene structure without mask supervision as well as surface rendering which acts as an inductive bias to represent objects by a set of precise 3D surfaces, leading to accurate reconstructions.

\subsection{Unifying Surface and Volume Rendering}

\noindent We start by noting that \eqnref{eq:nerf_original} can be rewritten as\footnote{We drop dependencies on the model parameters for clarity.}
\begin{equation}
\hat{C}(\br) = \sum_{i=1}^{N} \alpha_i(\bx_i) \prod_{j<i} \bigl(1-\alpha_j(\bx_j)\bigr) \, c(\bx_i, \bd)
\label{eq:nerf_alpha}
\end{equation}
with alpha values $\alpha_i(\bx)= 1 - \exp\left( - \sigma(\bx)\,\delta_i \right)$. Assuming solid objects, $\alpha$ becomes a discrete occupancy indicator variable $o\in\{0,1\}$ which either takes $o=0$ in free space and $o=1$ in occupied space as value:
\begin{equation}
\hat{C}(\br) = \sum_{i=1}^{N} o(\bx_i) \prod_{j<i} \bigl(1-o(\bx_j)\bigr) \, c(\bx_i, \bd)
\label{eq:nerf_occupancy}
\end{equation}
We recognize this expression as the image formation model for solid objects \cite{Ulusoy2015THREEDV} where the term $o(\bx_i) \prod_{j<i} \bigl(1-o(\bx_j)\bigr)$ evaluates to $1$ for the first occupied sample $\bx_i$ along ray $\br$ and to $0$ for all other samples. $\prod_{j<i} \bigl(1-o(\bx_j)\bigr)$ is an indicator for visibility which is $1$ if there exists no occupied sample $\bx_j$ with $j<i$ before sample $\bx_i$. Thus, $\hat{C}(\br)$ takes the color $c(\bx_i, \bd)$ of the first occupied sample along ray $\br$.

To unify implicit surface and volumetric radiance models, we parameterize $o$ directly by a continuos occupancy field $o_\theta$~\eqref{eq:onet} as opposed to predicting volume density $\sigma$.
Following \cite{Yariv2020ARXIV}, we condition the color field $c_\theta$ on the surface normal $\bn$ and a feature vector  $\bh$ of the geometry network which empirically induces a useful bias as also observed in \cite{Yariv2020ARXIV} for the case of implicit surfaces.
Importantly, our unified formulation allows for both volume and surface rendering
\begin{align}
\hat{C}_v(\br) &= \sum_{i=1}^{N} o_\theta(\bx_i) \prod_{j<i} \bigl(1-o_\theta(\bx_j)\bigr) c_\theta(\bx_i, \bn_i, \bh_i,\bd)
\label{eq:ours_volume}\\
\hat{C}_s(\br) &= c_\theta (\bx_s,\bn_s,\bh_s,\bd)
\label{eq:ours_surface}
\end{align}
where $\bx_s$ is retrieved by root-finding along ray $\br$ and $\bn_s$, $\bh_s$ denote the normal and geometry features at $\bx_s$, respectively. Note that $\bx_s$ depends on the occupancy field $o_\theta$, but we have dropped this dependency here for clarity. For further details, we refer the reader to the supplementary material.

The advantage of this unified formulation is that it allows for both rendering on the surface directly and rendering throughout the entire volume which enables gradually removing ambiguities during optimization.
As evidenced by our experiments, it is indeed critical to combine both for obtaining accurate reconstructions without mask supervision.
Being able to quickly recover the surface $\cS$ via root-finding enables more effective volume rendering, successively focusing on and refining the object surfaces as we will describe in \secref{sec:optimization}.
Furthermore, surface rendering enables faster novel view synthesis as illustrated in~\figref{fig:rendering methods}.

\subsection{Loss Function}

\noindent We optimize the following regularized loss function
\begin{equation}
\mathcal{L} = \cL_{rec} + \lambda\,\cL_{reg}
\label{eq:loss}
\end{equation}
with $\ell_1$ reconstruction loss and $\ell_2$ surface regularization which encourages the normal of a surface point $\bx_s$ and a point sampled in its neighborhood to be similar:
\begin{align}
\mathcal{L}_{rec} &= \sum_{\br \in \cR} {\lVert \hat{C}_v(\br) - C(\br) \rVert}_1 \label{eq:loss_rec}\\
\mathcal{L}_{reg} &= \sum_{\bx_s \in \cS} {\lVert \bn (\bx_s) - \bn (\bx_s+\bepsilon) \rVert}_2
\end{align}
Here, $\cR$ denotes the set of all pixels/rays in the minibatch, $\cS$ is the set of corresponding surface points, $C(\br)$ is the observed color for pixel/ray $\br$ and $\bepsilon$ is a small random uniform 3D perturbation.
The normal at $\bx_s$ is given by
\begin{equation}
\bn(\bx_s) = \frac{\nabla_{\bx_s} o_\theta(\bx_s)}{\lVert\nabla_{\bx_s} o_\theta(\bx_s)\rVert_2}
\end{equation}
which can be computed using double backpropagation~\cite{Niemeyer2020CVPR}.

\subsection{Optimization}
\label{sec:optimization}

The key hypothesis of implicit surface models \cite{Niemeyer2020CVPR,Yariv2020ARXIV} is that only the region at the first intersection point with the surface contributes to the rendering equation. However, this assumption is not true during early iterations where the surface is not well defined. Consequently, existing methods \cite{Niemeyer2020CVPR,Yariv2020ARXIV} require strong mask supervision. Conversely, during later iterations, knowledge of the approximate surface is valuable for drawing informative samples when evaluating the volume rendering equation in \eqnref{eq:ours_volume}.
Therefore, we utilize a training schedule with a monotonically decreasing sampling interval for drawing samples during volume rendering, as visualized in \figref{fig:concept}. In other words, during early iterations, the samples $\{\bx_i\}$ cover the entire optimization volume, effectively bootstrapping the reconstruction process using volume rendering. During later iterations, the samples $\{\bx_i\}$ are drawn closer around the estimated surface. As the surface can be estimated directly from the occupancy field $o_\theta$ via root-finding \cite{Niemeyer2020CVPR}, this eliminates the need for hierarchical two-stage sampling as in NeRF. Our experiments demonstrate that this procedure is particularly effective for estimating accurate geometry, while it allows for resolving ambiguities during early iterations.

More formally, let $\bx_s=\bo + t_s \bd$. We obtain samples $\bx_i=\bo + t_i \bd$ by drawing $N$ depth values $t_i$ using stratified sampling within the interval $[t_s-\Delta, t_s+\Delta]$ centered at $t_s$:
\begin{equation}
t_i \sim \mathcal{U}\left[ t_s +\left( \frac{2i-2}{N}-1\right) \Delta, t_s +\left( \frac{2i}{N}-1\right) \Delta \right]
\end{equation}  
During training, we start with a large interval $\Delta_\text{max}$ and gradually decrease $\Delta$ for more accurate sampling and optimization of the surface using the following decay schedule
\begin{equation}
\Delta_k = \max(\Delta_\text{max} \exp (-k \, \beta), \Delta_\text{min})
\label{eq:adaptive}
\end{equation}
where $k$ denotes the iteration number and $\beta$ is a hyperparameter.
In fact, it can be shown that for $\Delta \rightarrow 0$ and $N \rightarrow \infty$, volume rendering \eqref{eq:ours_volume} indeed approaches surface rendering \eqref{eq:ours_surface}: $\hat{C}_v(\br) \rightarrow \hat{C}_s(\br)$. A formal proof of this limit is provided in the supplementary material.

As evidenced by our experiments, the decay schedule in \eqref{eq:adaptive} is critical for capturing detailed geometry as it combines volume rendering of large and uncertain volumes in the beginning of training with surface rendering towards the end of training.
To reduce free space artifacts,  we combine these samples with points sampled randomly between the camera and the surface.
For rays without surface intersection, we use stratified sampling on the entire ray.

\subsection{Implementation Details}

\boldparagraph{Architecture}
Similar to Yariv \etal \cite{Yariv2020ARXIV}, we use an 8-layer MLP with a Softplus activation function and a hidden dimension of $256$ for the occupancy field $o_\theta$. We initialize the network such that the decision boundary is a sphere \cite{Gropp2020ICML}. In contrast, the radiance field $c_\theta$ is parameterized as a 4-layer ReLU MLP.
We encode the 3D location~$\bx$ and the viewing direction~$\bd$ using Fourier features \cite{Mildenhall2020ECCV} at $k$ octaves.
We empirically found $k=6$ for the 3D location $\bx$ and $k=4$ for the viewing direction $\bd$ to work best.

\boldparagraph{Optimization}
In all experiments, we fit our model to multi-view images of a single scene.
During optimization of the model parameters, we first randomly sample a view and then $M$ pixels/rays $\cR$ from this view based on the camera intrinsics and extrinsics.
Next, we render all rays to compute the loss function in \eqnref{eq:loss}.
For root-finding, we use $256$ uniform sampled points and apply the secant method with $8$ steps \cite{Mescheder2019CVPR}.
For our rendering procedure, we use $N=64$ query points inside the interval and $32$ in the free space between the camera and the lower bound of the interval.
The interval decay parameters are $\beta=1.5e-5$, $\Delta_\text{min}=0.05$ and $\Delta_\text{max}=1.0$.
We use Adam with a learning rate of $0.0001$ and optimize for $M=1024$ pixels per iteration with two decay steps after $200$k and $400$k iterations. In total, we train our models for $450$k iterations.

\begin{figure}
	\center
	\begin{tabular}{@{}c@{ }c@{ }c}
		\includegraphics[width=0.33\linewidth]{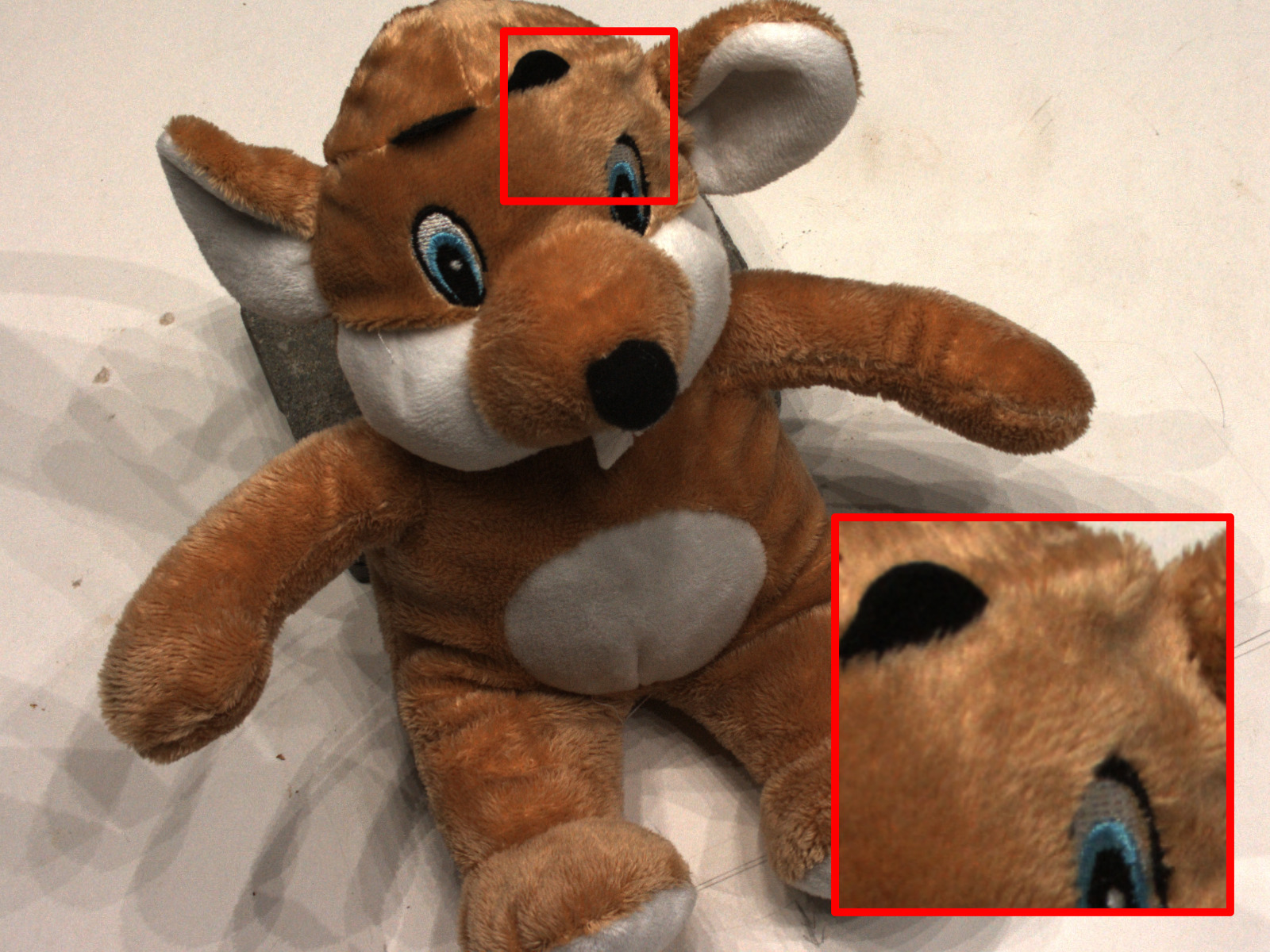}&
		\includegraphics[width=0.33\linewidth]{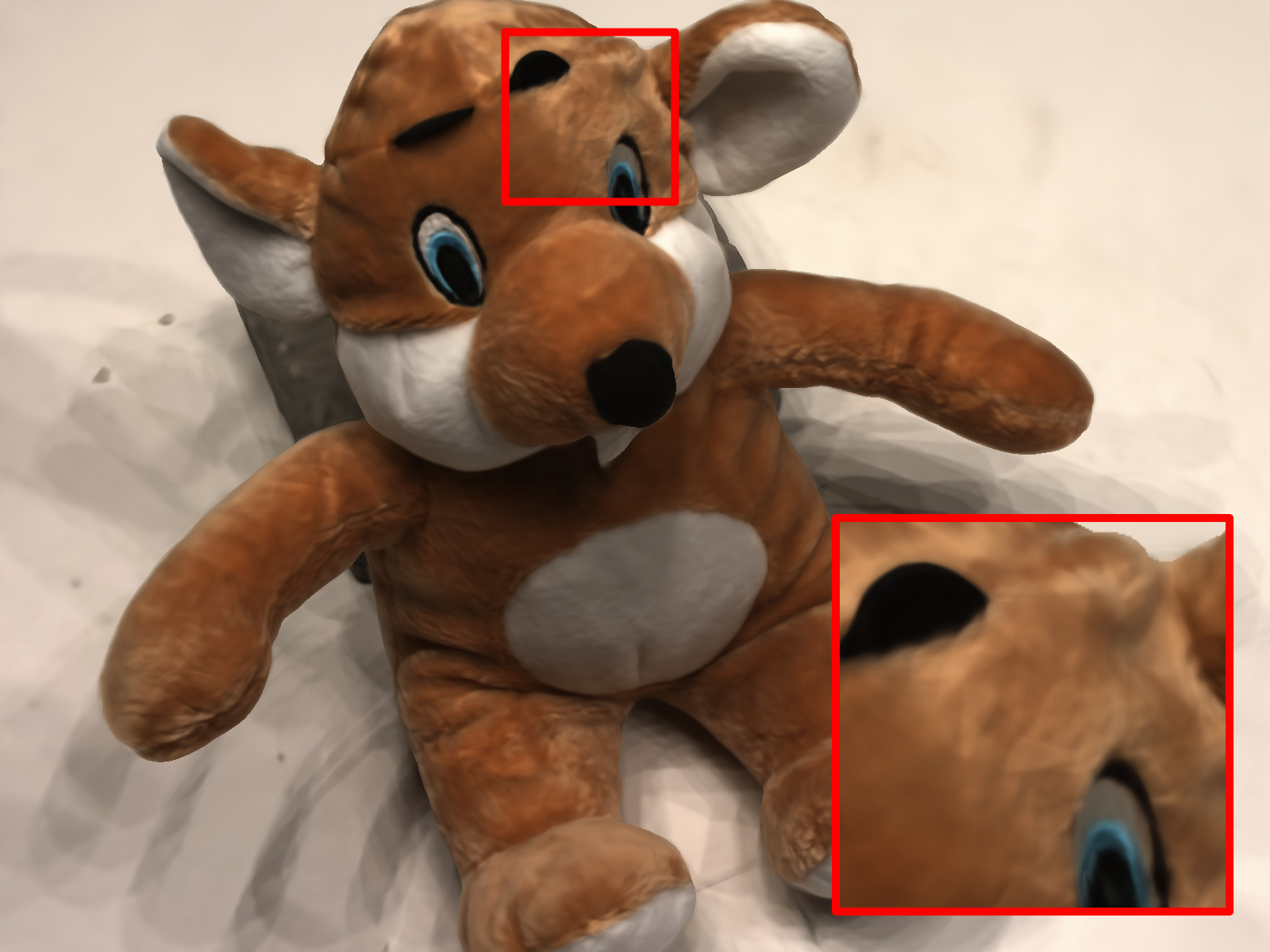}&
		\includegraphics[width=0.33\linewidth]{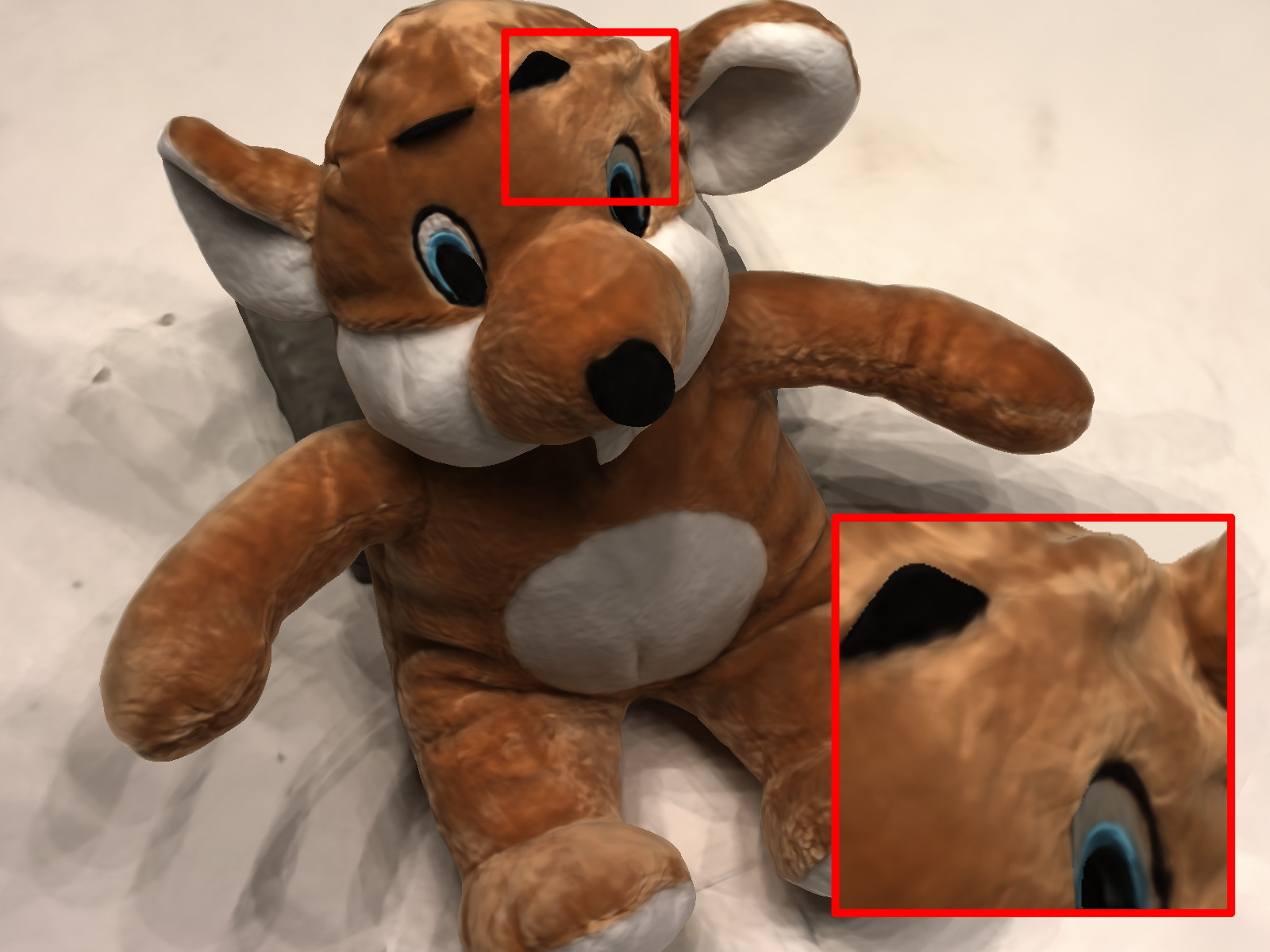}\\
		GT&
		VR ($178\,\text{s}$)\footnotemark&
		SR ($84\,\text{s}$)\\
		&
		$32/0.79/0.33$&
		$30/0.78/0.32$\\
	\end{tabular}
	\vspace{-0.2cm}
	\caption{\textbf{Volume \vs Surface Rendering.} We compare images generated using volume rendering (VR, $\Delta = \Delta_\text{min}$) and surface rendering (SR), reporting three image metrics (PSNR$\uparrow$/SSIM$\uparrow$/LPIPS$\downarrow$). Both approaches yield similar results while surface rendering is twice as fast.} %
	\label{fig:rendering methods}
	\vspace{-0.2cm}
\end{figure}
\footnotetext{We used a single NVIDIA V100 GPU to measure the inference time.}

\boldparagraph{Inference}
Our method allows for inferring 3D shapes as well as for synthesizing novel view images.
For synthesizing images, we can render our representation in two different ways, we can either use volume rendering or surface rendering.
In \figref{fig:rendering methods}, we show that both rendering approaches lead to similar results. However, we observe that surface rendering is faster than volume rendering.

To extract meshes, we apply the Multiresolution IsoSurface Extraction (MISE) algorithm from \cite{Mescheder2019CVPR}. We use $64^3$ as the initial resolution and up-sample the mesh in 3 steps without gradient-based refinement. 

%% file: sec_results.tex
\section{Experimental Evaluation}
\label{sec:results}

We conduct experiments on multi-view 3D reconstruction to assess our method.
First, we provide a qualitative and quantitative comparison of our approach to existing methods (IDR~\cite{Yariv2020ARXIV}, NeRF~\cite{Mildenhall2020ECCV}, COLMAP~\cite{Schoenberger2016ECCV}) on the widely used \emph{DTU MVS} dataset \cite{Jensen2014CVPR}.
Second, we show a qualitative comparison for samples from the \emph{BlendedMVS} dataset~\cite{Yao2020CVPR} and synthetic renderings of scenes from the \emph{SceneNet} dataset~\cite{McCormac2017ICCV}.
Third, we analyze our rendering procedure and loss function in an ablation study.
In the supplementary, we provide results on the LLFF dataset \cite{Mildenhall2019SIGGRAPH}.

\subsection{Baselines}
 To validate the effectiveness of our method, we compare it to three different baselines.

 \boldparagraph{COLMAP~\cite{Schoenberger2016ECCV}}
We consider COLMAP~\cite{Schoenberger2016ECCV} as a classical MVS baseline, as it shows strong performance on multi-view reconstruction and is widely used in related works \cite{Niemeyer2020CVPR, Yariv2020ARXIV}.
 We reconstruct a mesh from the output of COLMAP using screened Poisson Surface reconstruction (sPSR) \cite{Kazhdan2013SIGGRAPH}.
 Following \cite{Niemeyer2020CVPR,Yariv2020ARXIV}, we show results for the quantitatively best trim parameter $7$ and for the setting that results in watertight meshes (trim parameter $0$).

 \boldparagraph{NeRF~\cite{Mildenhall2020ECCV}}
 Although NeRF targets novel view synthesis, its volume density admits the extraction of geometry. To extract meshes from NeRF, we define a density threshold of 50. We validate this choice in the supplementary.

 \boldparagraph{IDR~\cite{Yariv2020ARXIV}}
 IDR is the state-of-the-art multi-view reconstruction method for neural implicit surfaces. 
 IDR reconstructs surfaces with an impressively high level of detail and handles specular surfaces, but requires input masks. We do not compare to DVR~\cite{Niemeyer2020CVPR} as IDR's view-dependent modeling has been demonstrated to be superior to DVR, see \cite{Yariv2020ARXIV}.

\subsection{Datasets}
\vspace{-0.3cm}
\boldparagraph{\emph{DTU MVS} Dataset~\cite{Jensen2014CVPR}} 
The \emph{DTU MVS} dataset contains 49 to 64 images at a resolution of $1200\times 1600$ as well as extrinsic and intrinsic camera parameters for all views. 
The dataset consists of objects with different shapes and appearances. Non-lambertian appearance effects make some of the objects particularly challenging.
For each scan, ground truth 3D shapes, as well as the official evaluation procedure, are available\footnote{\url{https://roboimagedata.compute.dtu.dk}}. 
Like previous works \cite{Niemeyer2020CVPR, Yariv2020ARXIV}, we use the "Surface" method of the evaluation script, and evaluate all methods on meshes cleaned with the respective masks. 
The official evaluation procedure calculates the Chamfer distance between sampled points of the predicted shapes and ground truth shapes provided in the dataset.
For evaluating IDR \cite{Yariv2020ARXIV}, we use the pixel-accurate masks for all images which are provided by the authors of IDR.

\boldparagraph{\emph{BlendedMVS} Dataset~\cite{Yao2020CVPR}}
The \emph{BlendedMVS} dataset is a large-scale dataset containing multi-view images with respective camera extrinsics and intrinsics. We use examples from the \emph{BlendedMVS} low-res set with an image resolution of $768\times576$.
These examples contain 24 to 64 different views of unmasked images.
We define the scene as such that the object is in the center and the closest camera lies near the unit sphere.

\boldparagraph{\emph{SceneNet} Dataset~\cite{McCormac2017ICCV}}
For testing our model on complex indoor scenes, we take two scenes from \cite{McCormac2017ICCV} for evaluation.
\emph{BlenderProc} \cite{Denninger2019Arxiv} is applied for rendering images of a part of the scenes containing multiple objects.
The first scene is a bedroom scene with a bed, a lamp and a bedside table. The other scene is a living room scene with a sofa, a curtain and a round table. We use 83 and 40  images, respectively.

\subsection{Comparison on DTU}
 \begin{table}[h!]
	 \center
    \resizebox{\linewidth}{!}{
	 \input{tab/table_3drec_nosr.tex}}
	 \caption{\textbf{Quantitative Comparison on DTU~\cite{Jensen2014CVPR}.} We show a quantitative comparison against the baselines for the reconstructed geometry on 15 scans from the DTU dataset. Our method performs on par with IDR, even though IDR uses masks as input. Bold numbers are only considering methods without mask supervision and trimming.}
	 \label{tab:3drec}
   \vspace{-0.5cm}
 \end{table}

 In \tabref{tab:3drec}, we quantitatively compare our method to the baselines on the \emph{DTU MVS} dataset.
 While the COLMAP baseline with trim parameter $\zeta = 7$ shows the best performance on the Chamfer distance, it produces non-watertight meshes with incomplete surfaces.
 Our method performs nearly on par with the state-of-the-art neural implicit model IDR while not relying on strong mask supervision.
 NeRF and COLMAP ($\zeta = 0$) also do not use input masks but exhibit worse performance in terms of Chamfer distance. 

\begin{figure*}[h]
	 \center
	 \input{tab/fig_3drecwomasks.tex}
	 \caption{\textbf{Reconstructed 3D Shapes for DTU.} We show a qualitative comparison of reconstructed surfaces. We use COLMAP~\cite{Schoenberger2016ECCV}, NeRF~\cite{Mildenhall2020ECCV} and IDR~\cite{Yariv2020ARXIV} as baselines. While IDR requires pixel-accurate object masks for all images, the other methods reason about the full scene extent. Our method performs visually on par with IDR while not requiring masks.}
	 \label{fig:qulicomp_dtumasks}
   \vspace{-0.3cm}
 \end{figure*}
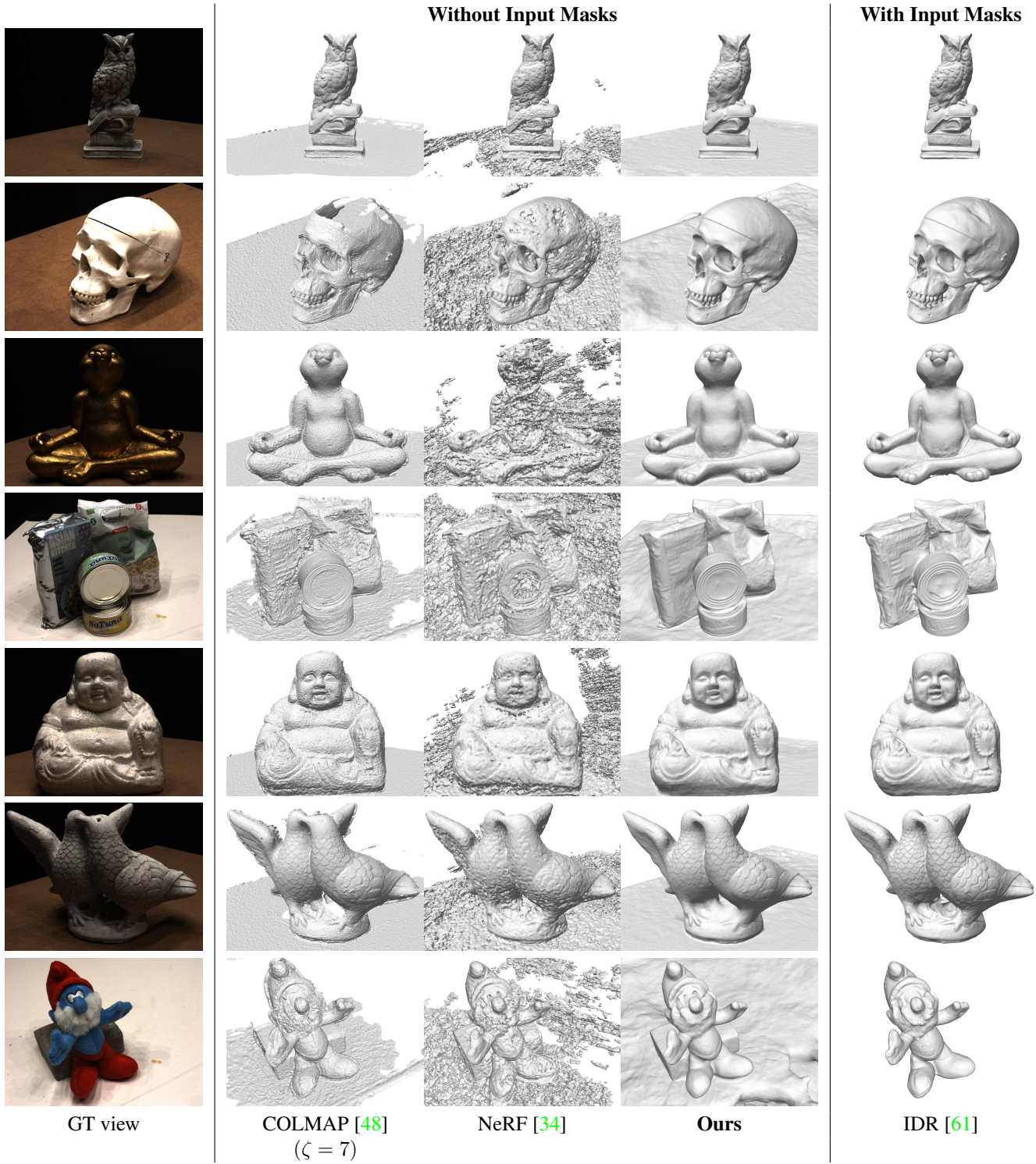

In \figref{fig:qulicomp_dtumasks}, we show qualitative results for our method and the baselines.
While COLMAP provides detailed reconstructions, it leads to incomplete geometry due to trimming.
For NeRF, holes and noise artifacts can be observed in the reconstructions.
In contrast, our approach and IDR (with input masks) produce accurate surfaces with high-quality details.
We remark that our model captures the overall spatial arrangement of the scene accurately while being able to also capture geometric details, \eg, the teeth of the skull and other surface details.

 \begin{figure*}[h!]
   \center
   \begin{tabular}{c@{}c@{}c@{}c@{}c@{}c}
      \includegraphics[width=0.16\linewidth]{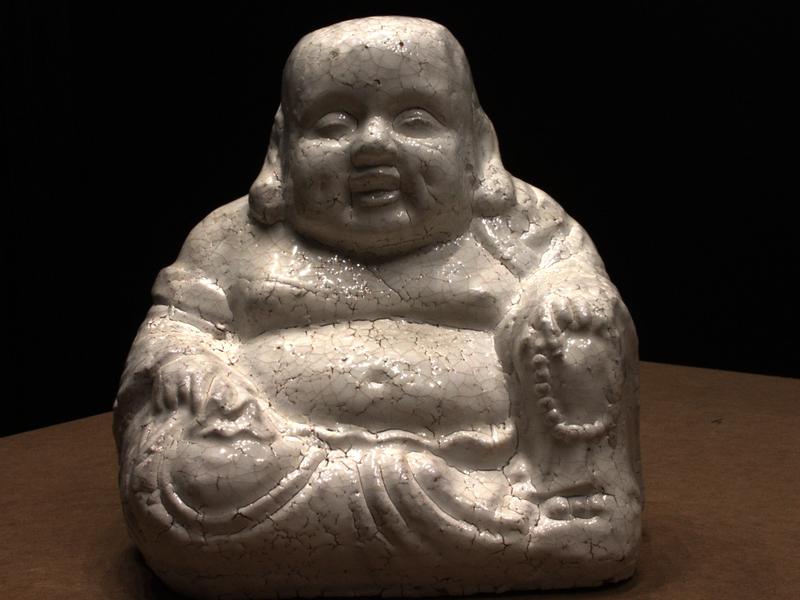}&
      \includegraphics[width=0.16\linewidth]{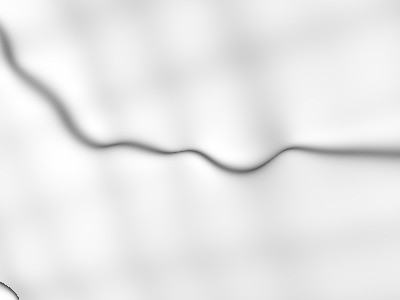}&
      \includegraphics[width=0.16\linewidth]{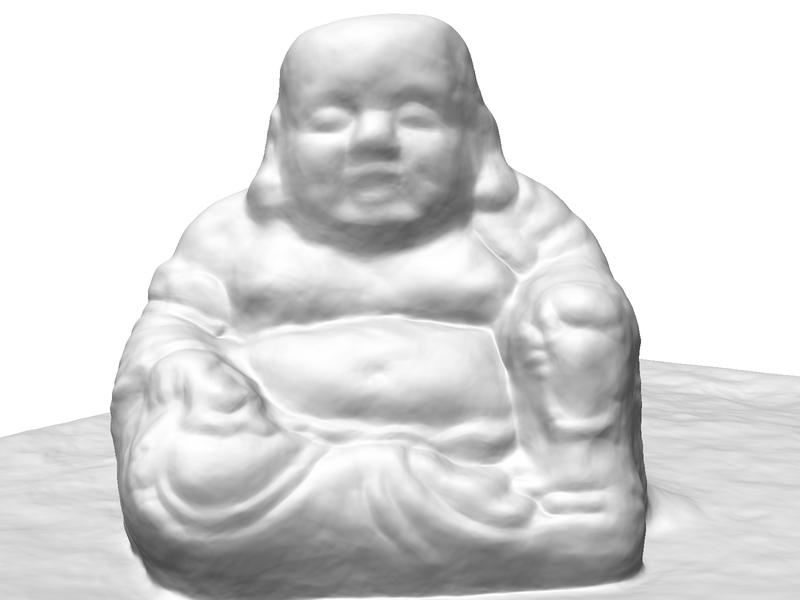}&
      \includegraphics[width=0.16\linewidth]{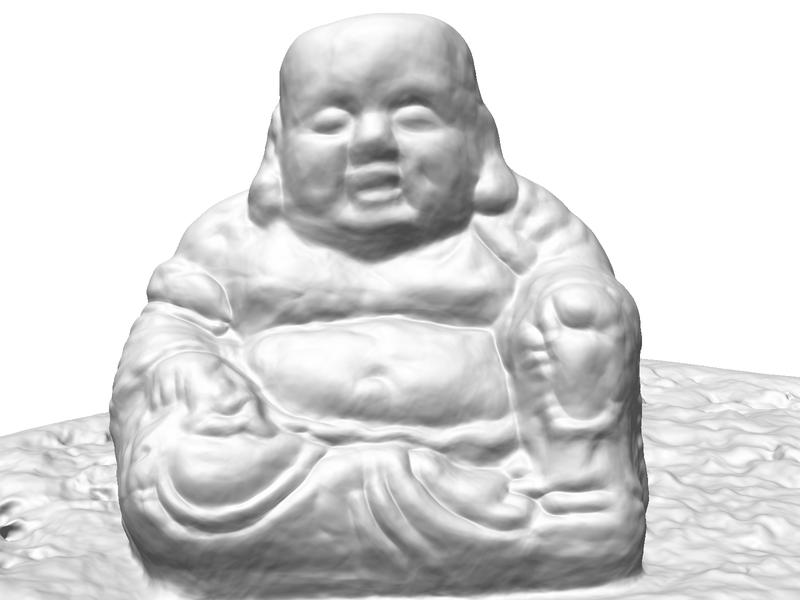}&
      \includegraphics[width=0.16\linewidth]{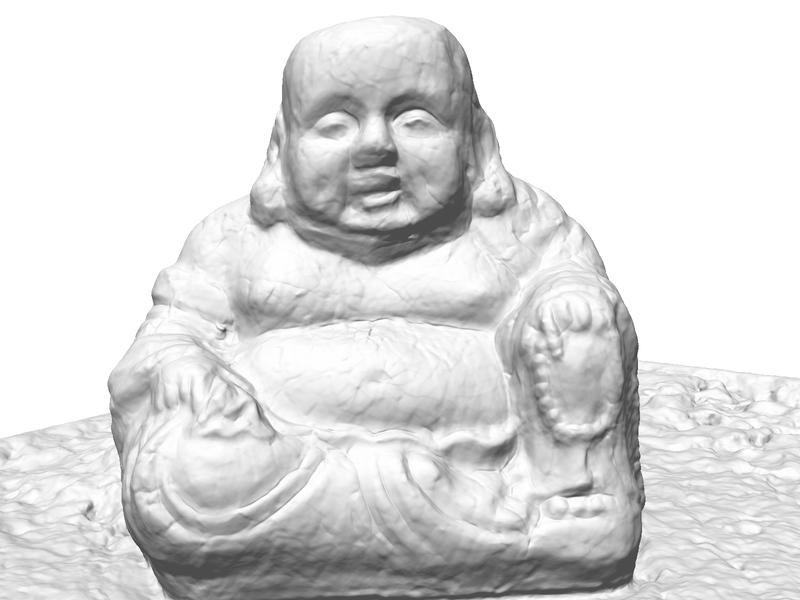}&
      \includegraphics[width=0.16\linewidth]{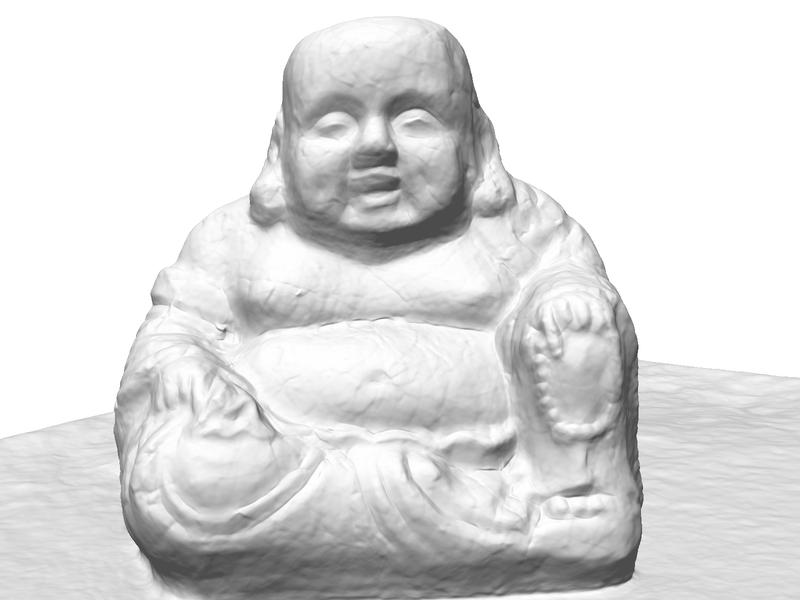}\\
      &
      SR&
      Uniform VR&
     HVR&
     Ours w/o Surf Reg&
     Ours \\
      Chamfer Distance&
      (not converging)&
      3.70&
     3.22&
     0.48&
     0.46\\
   \end{tabular}
   \vspace{-0.3cm}
   \caption{\textbf{Ablation Study.} We compare our method to different rendering procedures: Surface rendering in isolation (SR), volume rendering with uniformly sampled points (Uniform VR) and hierachical volume rendering (HVR). This analysis on the Buddha scene (DTU) shows that our full method performs best, both visually and quantitatively.
   Moreover, our model without surface regularizer (Ours w/o Surf Reg) leads to non-smooth regions, especially at the textureless flat table.}
   \label{fig:rendering_ablations}
   \vspace{-0.3cm}
\end{figure*}

 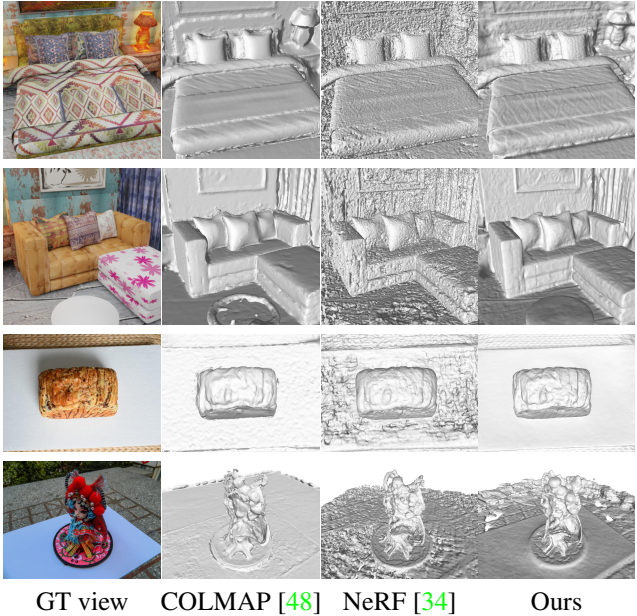
\begin{figure}[h]
   \center
   \input{tab/fig_3drec_indoor.tex}
   \caption{\textbf{Reconstructions for Indoor Scenes~\cite{McCormac2017ICCV} and BlendedMVS~\cite{Yao2020CVPR}.} We show a qualitative comparison of reconstructed surfaces on indoor scenes (1st and 2nd row) and \textit{BlendedMVS} (3rd, 4th and 5th row).}
   \label{fig:qulicomp_indoor}
   \vspace{-0.3cm}
\end{figure}

 \subsection{Comparison on BlendedMVS and SceneNet}
 
To show our model's capabilities on more diverse scenes, we use samples of the \textit{BlendedMVS} dataset and indoor scenes from \textit{SceneNet}. 
As there exist no object masks for these scenes, we only consider COLMAP and NeRF as baseline methods.
While we have tried running IDR, none of the scenes converged, resulting in degenerate outputs.
For scenes with complex backgrounds, we use a background model that learns to capture appearance information outside of our region interest, we refer the reader to the supplementary material for more details.

Our qualitative results in \figref{fig:qulicomp_indoor} provide evidence that, unlike existing implicit surface models, our method is able to reconstruct plausible geometry for complex scenes with multiple objects and backgrounds.
While COLMAP works well on indoor scenes, it shows artifacts on uniformly colored regions (\eg, round table in the second row).  
As for the \textit{BlendedMVS} experiment, NeRF can reason about the overall spatial structure but shows less accurate surfaces with significantly higher levels of noise compared to UNISURF.
More results can be found in the supplementary material.

\subsection{Ablation study}

We finally investigate the impact of rendering design choices and show ablations of the loss function.

\boldparagraph{Rendering Procedure}
We argue for our choices of the rendering procedure by comparing different variations in \figref{fig:rendering_ablations}:
First, we consider a baseline which only uses surface rendering during optimization (SR).
We use an $\ell_1$ reconstruction loss on the surface color and backpropagate through implicit differentiation following \cite{Niemeyer2020CVPR}.
A second baseline uses uniform volume rendering with 96 query points (Uniform VR).
Third, we use NeRF's hierarchical volume sampling (HVR) \cite{Mildenhall2020ECCV} with 64+64 sample points for querying the occupancy field.
While surface rendering does not converge, Uniform VR and HVR result in overly smooth and bloated shapes with missing details.
This provides evidence that the proposed unified model leads to more accurate reconstructions compared to the baselines.

\boldparagraph{Losses}
In \figref{fig:rendering_ablations}, we also show an ablation study of our surface regularization term in \eqref{eq:loss}.
Without this regularizer surfaces become less smooth in flat and ambiguous areas, \eg, at the table.
This regularization term is particularly useful for regions that are observed less frequently, as it incorporates an inductive bias towards smooth surfaces.

%% file: tab/table_3drec_nosr.tex
\begin{tabular}{l|c|c|ccc}
\toprule
{} &        COLMAP &            IDR &        COLMAP &  NeRF &           Ours \\
masks &     \xmark &         \cmark &         \xmark & \xmark & \xmark \\
trim &         7  &          - &  0 &  -& -\\
\midrule
scan24  &  0.45 &  1.63 &  \textbf{0.81} &  1.90 &           1.32 \\
scan37  &  0.91 &  1.87 &           2.05 &  1.60 &  \textbf{1.36} \\
scan40  &  0.37 &  0.63 &  \textbf{0.73} &  1.85 &           1.72 \\
scan55  &  0.37 &  0.48 &           1.22 &  0.58 &  \textbf{0.44} \\
scan63  &  0.90 &  1.04 &           1.79 &  2.28 &  \textbf{1.35} \\
scan65  &  1.00 &  0.79 &           1.58 &  1.27 &  \textbf{0.79} \\
scan69  &  0.54 &  0.77 &           1.02 &  1.47 &  \textbf{0.80} \\
scan83  &  1.22 &  1.33 &           3.05 &  1.67 &  \textbf{1.49} \\
scan97  &  1.08 &  1.16 &           1.40 &  2.05 &  \textbf{1.37} \\
scan105 &  0.64 &  0.76 &           2.05 &  1.07 &  \textbf{0.89} \\
scan106 &  0.48 &  0.67 &           1.00 &  0.88 &  \textbf{0.59} \\
scan110 &  0.59 &  0.90 &  \textbf{1.32} &  2.53 &           1.47 \\
scan114 &  0.32 &  0.42 &           0.49 &  1.06 &  \textbf{0.46} \\
scan118 &  0.45 &  0.51 &           0.78 &  1.15 &  \textbf{0.59} \\
scan122 &  0.43 &  0.53 &           1.17 &  0.96 &  \textbf{0.62} \\
\midrule
mean    &  0.65 &  0.90 &           1.36 &  1.49 &  \textbf{1.02} \\
\bottomrule
\end{tabular}

%% file: tab/fig_3drecwomasks.tex
\newcommand{\mywidth}{0.192\linewidth}
\begin{tabular}{@{}c|c@{}c@{}c|c}
    {} &{}&\textbf{Without Input Masks} &{}&{\textbf{With Input Masks}}\\
    \includegraphics[width=\mywidth]{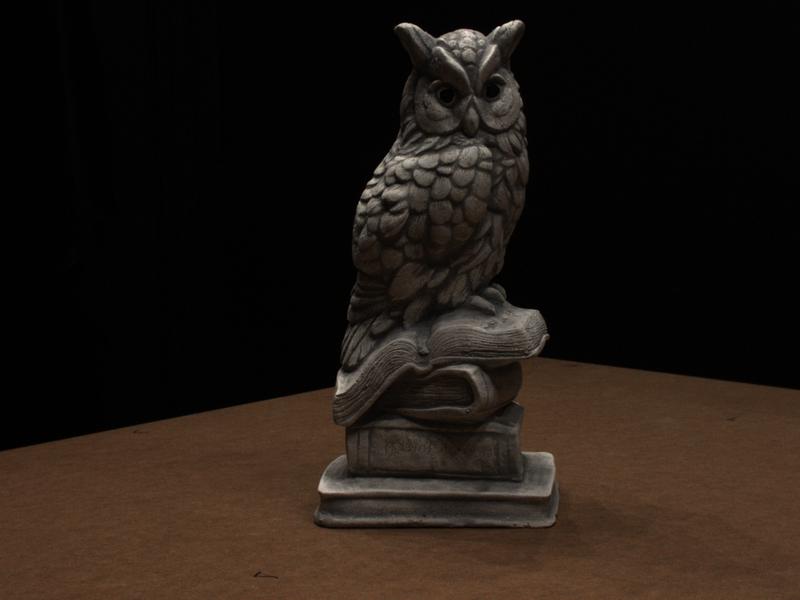}&
    \includegraphics[width=\mywidth]{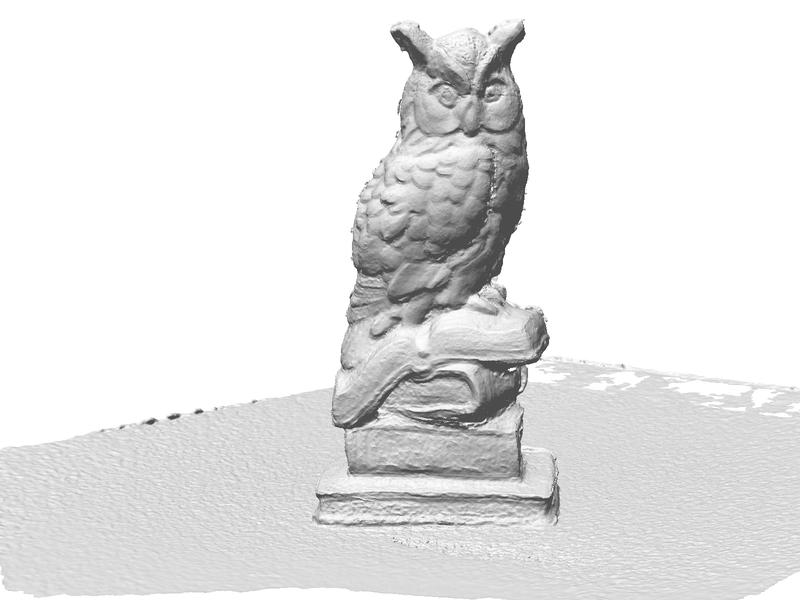}&
    \includegraphics[width=\mywidth]{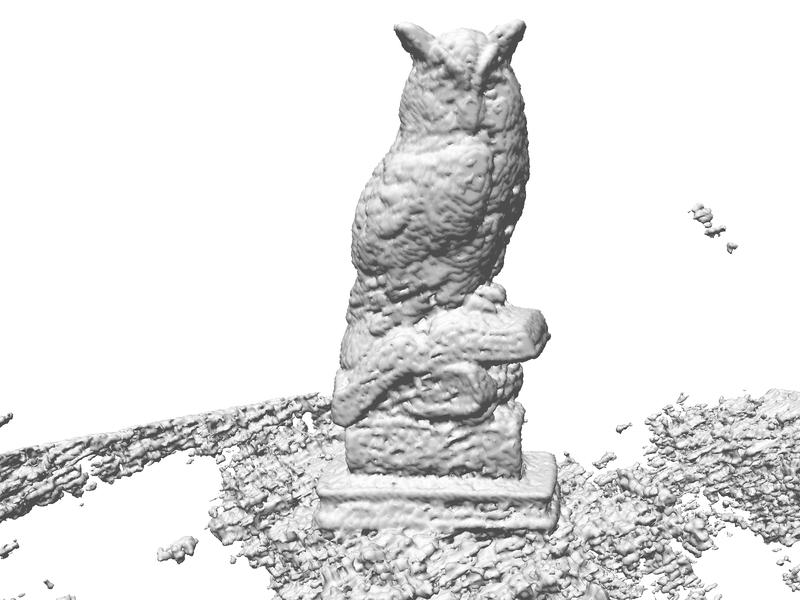}&
    \includegraphics[width=\mywidth]{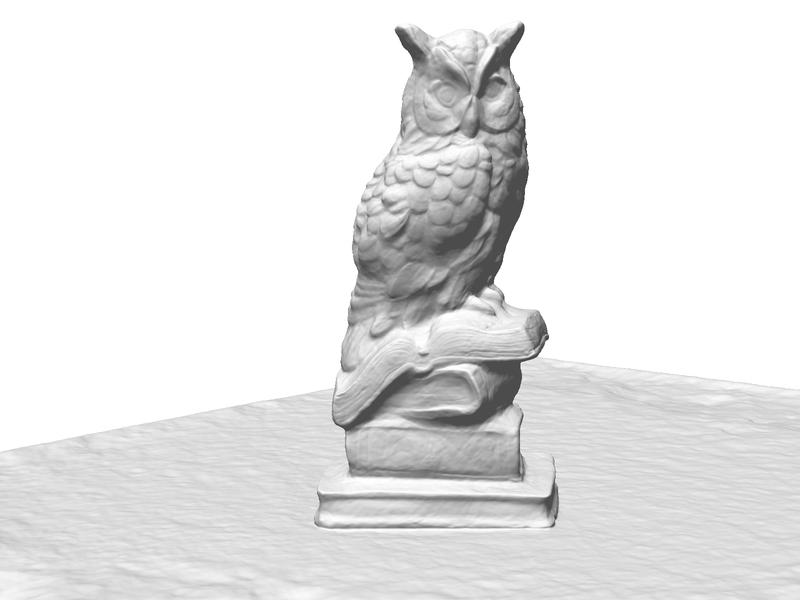}&
    \includegraphics[width=\mywidth]{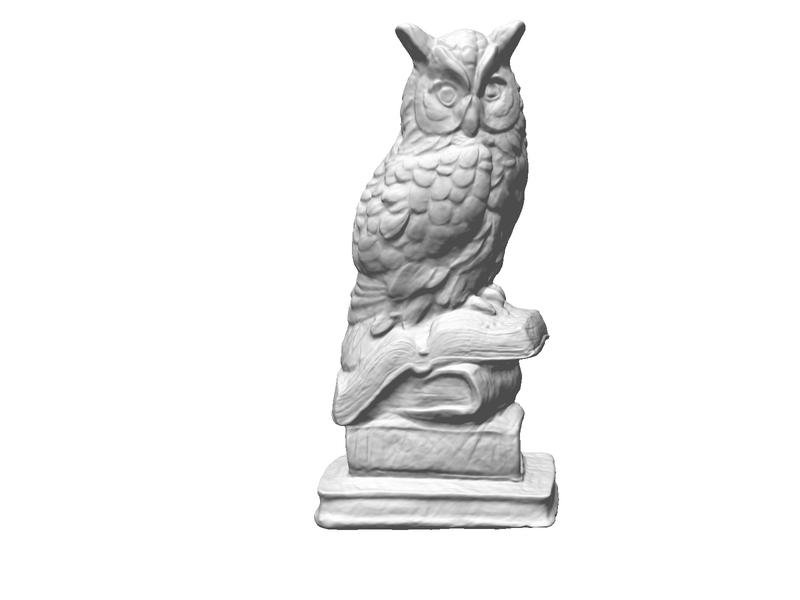}\\
    \includegraphics[width=\mywidth]{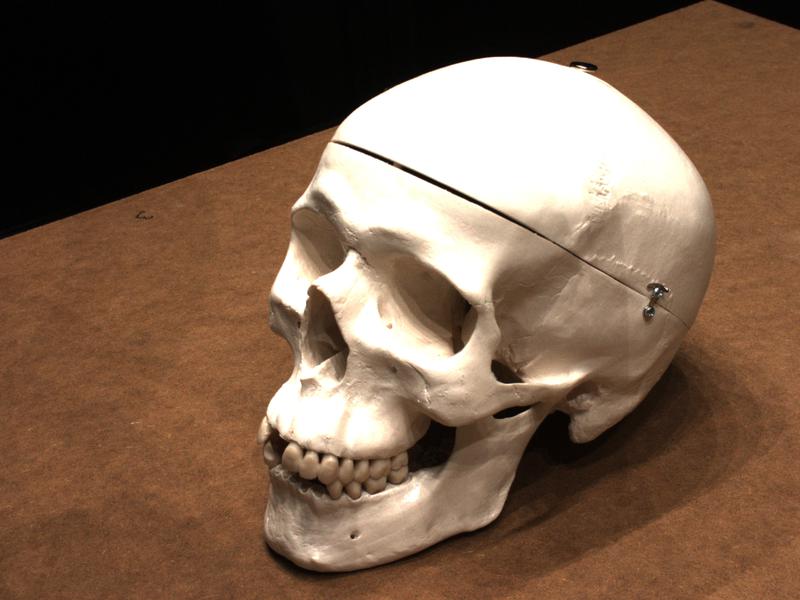}&
    \includegraphics[width=\mywidth]{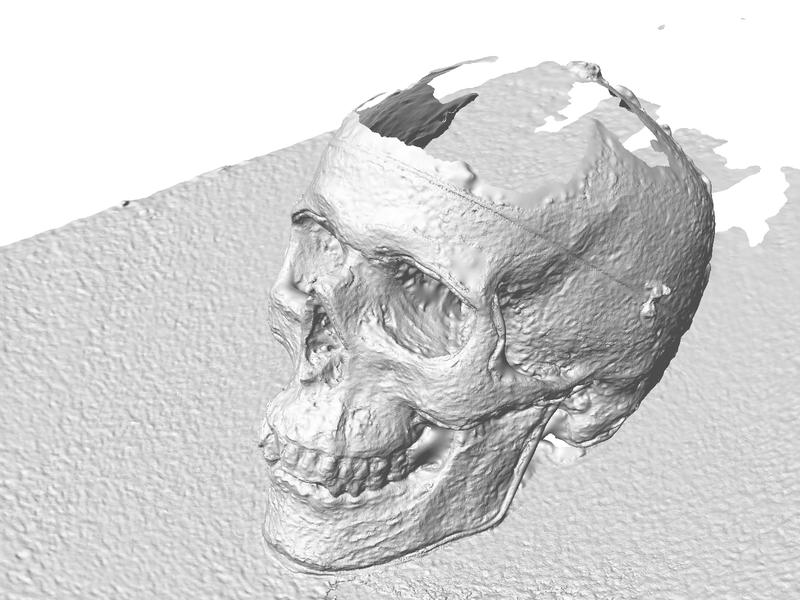}&
    \includegraphics[width=\mywidth]{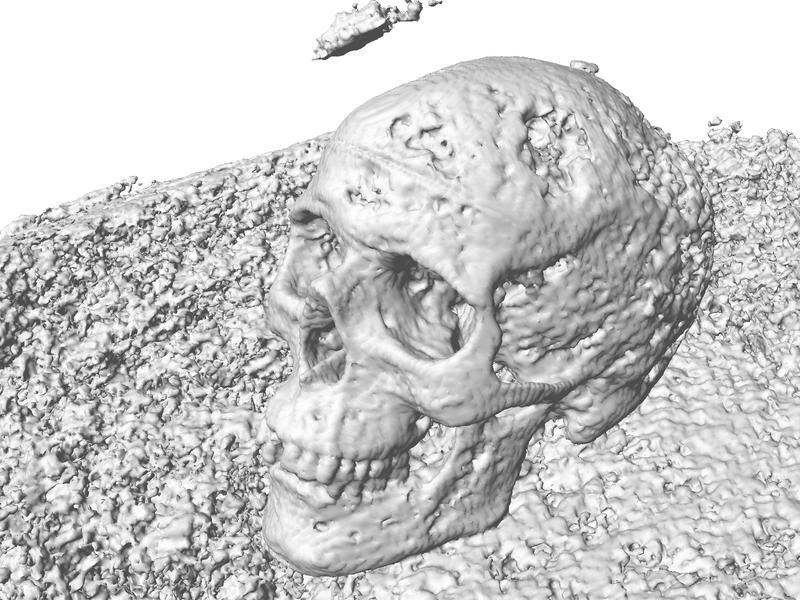}&
    \includegraphics[width=\mywidth]{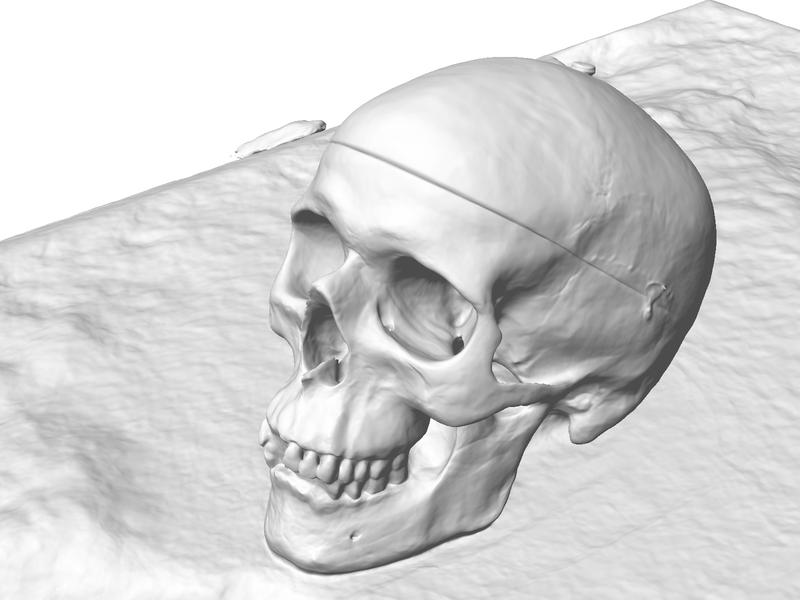}&
    \includegraphics[width=\mywidth]{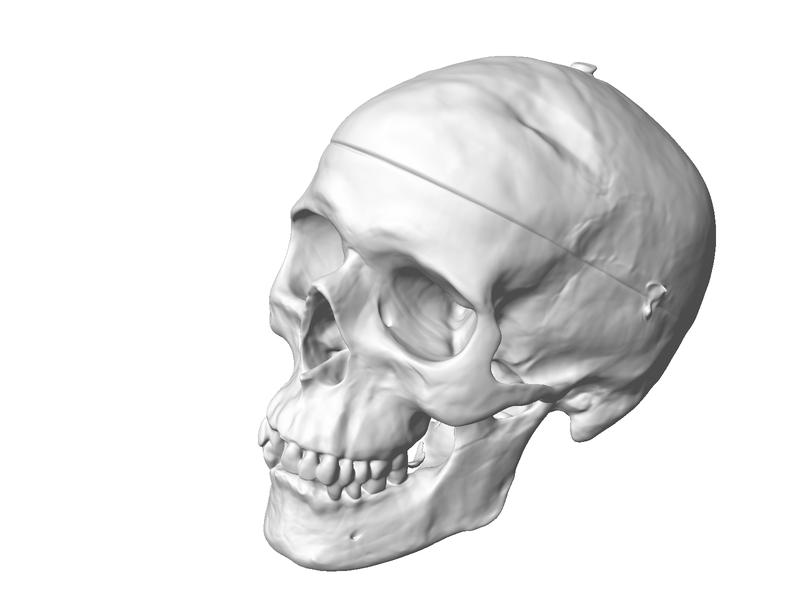}\\
    \includegraphics[width=\mywidth]{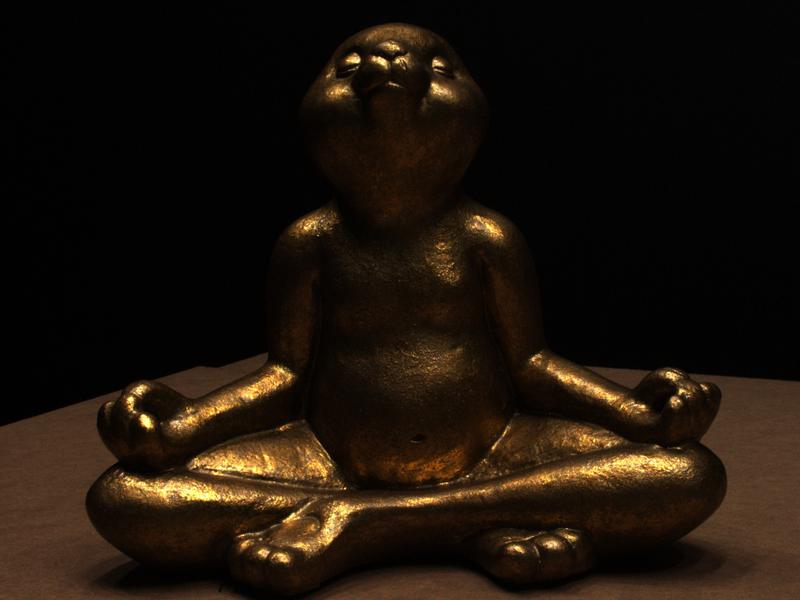}&
    \includegraphics[width=\mywidth]{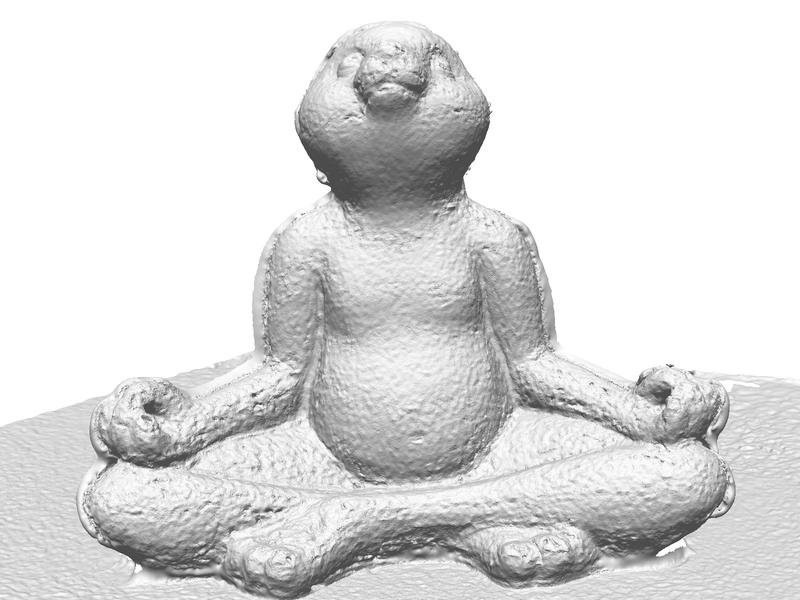}&
    \includegraphics[width=\mywidth]{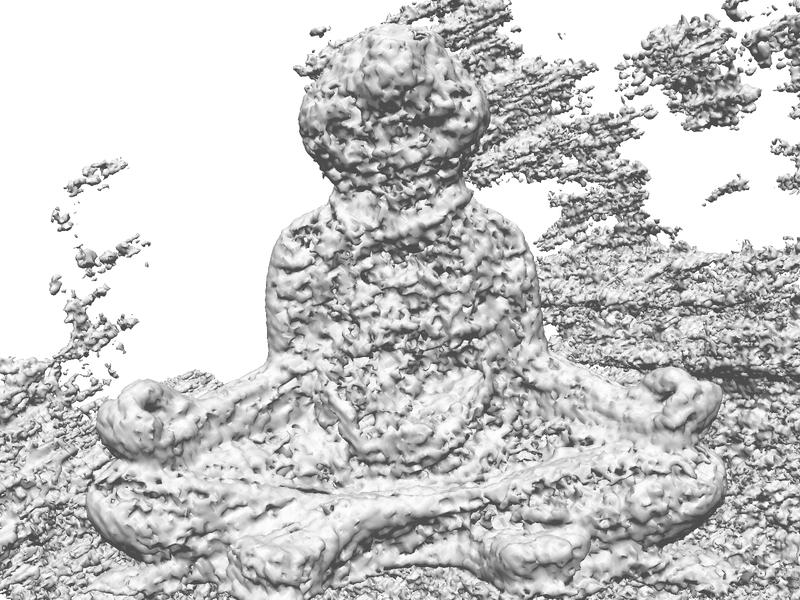}&
    \includegraphics[width=\mywidth]{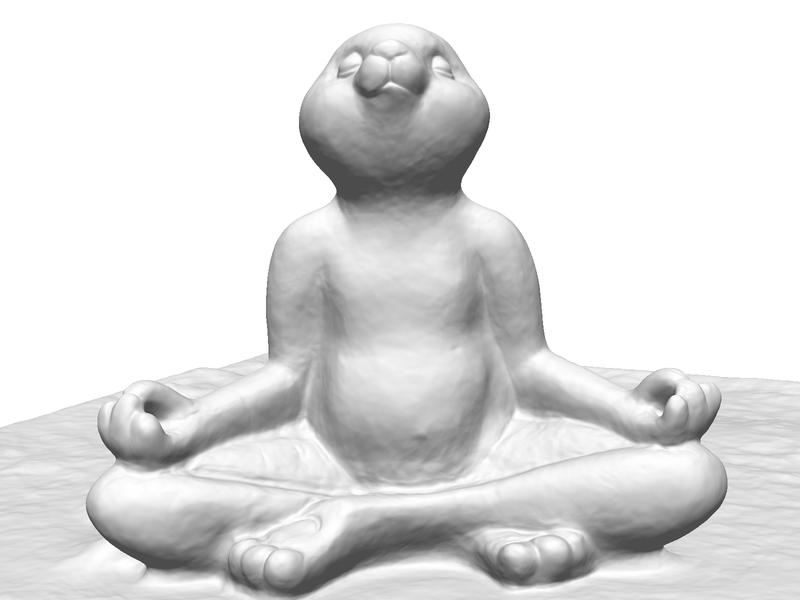}&
    \includegraphics[width=\mywidth]{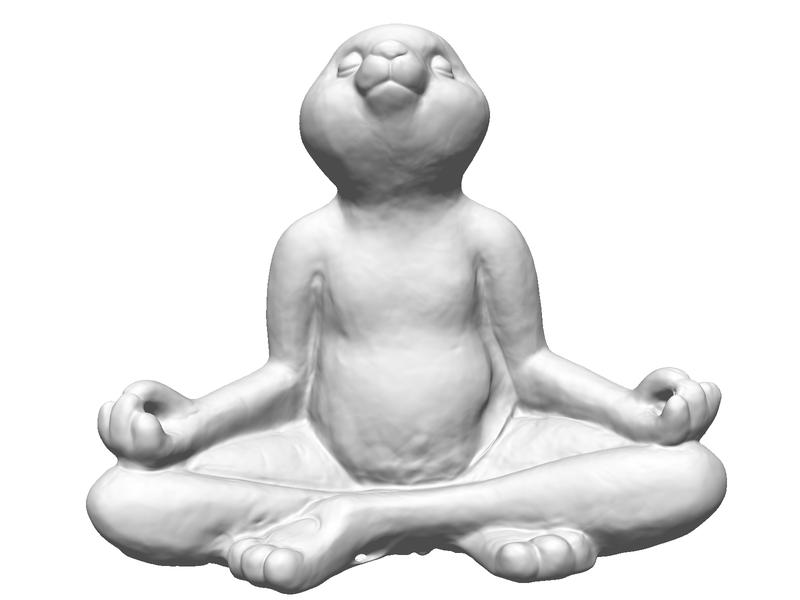}\\
    \includegraphics[width=\mywidth]{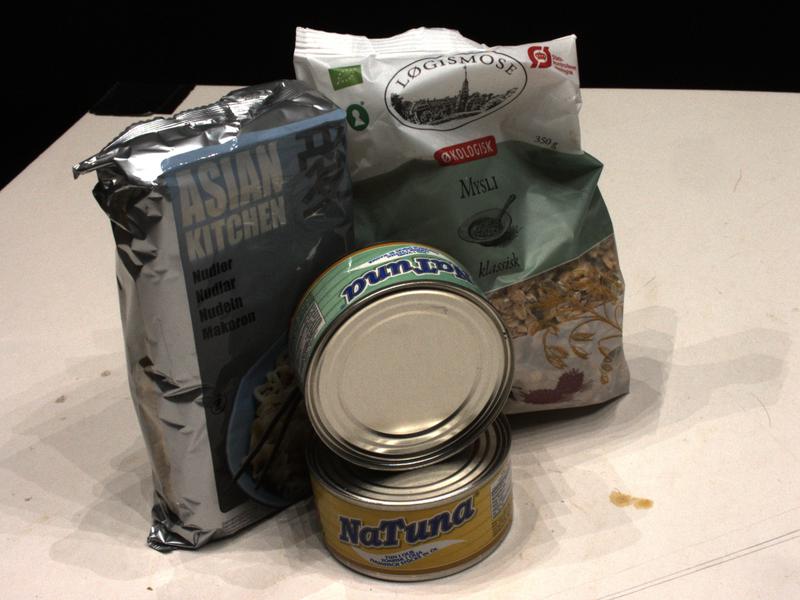}&
    \includegraphics[width=\mywidth]{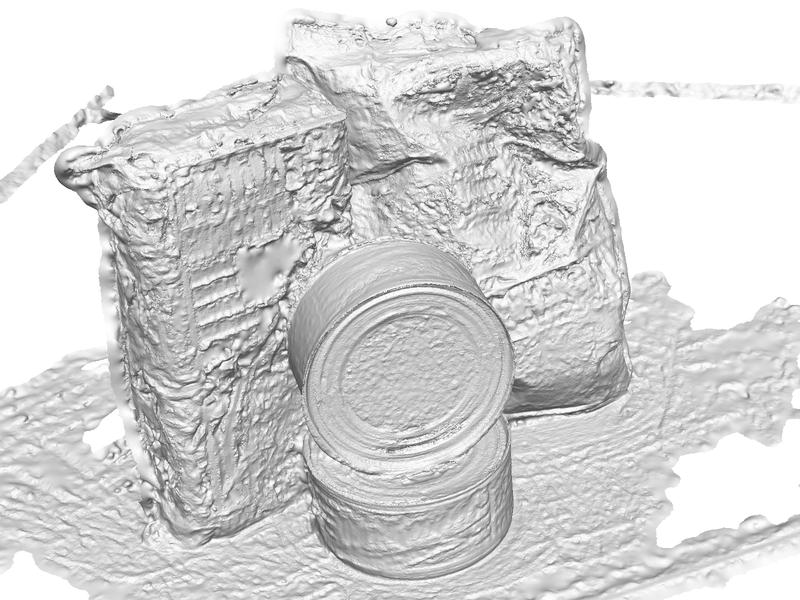}&
    \includegraphics[width=\mywidth]{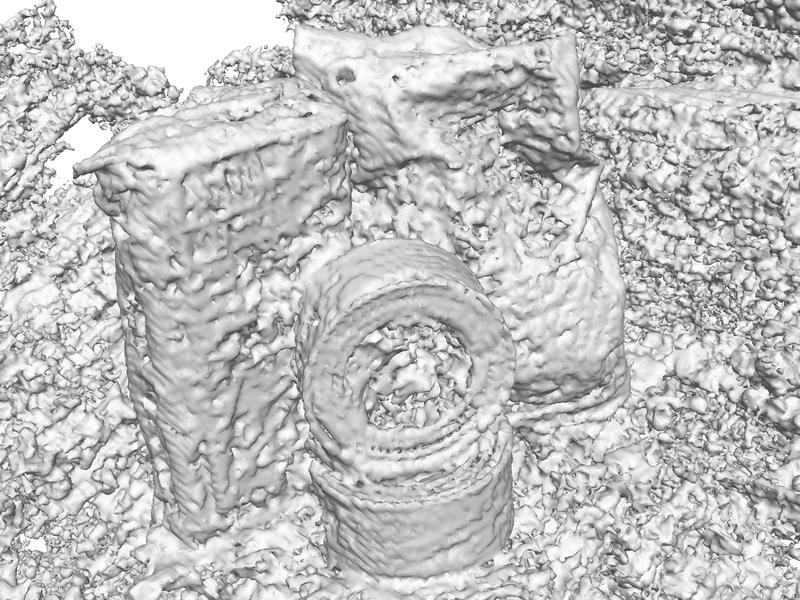}&
    \includegraphics[width=\mywidth]{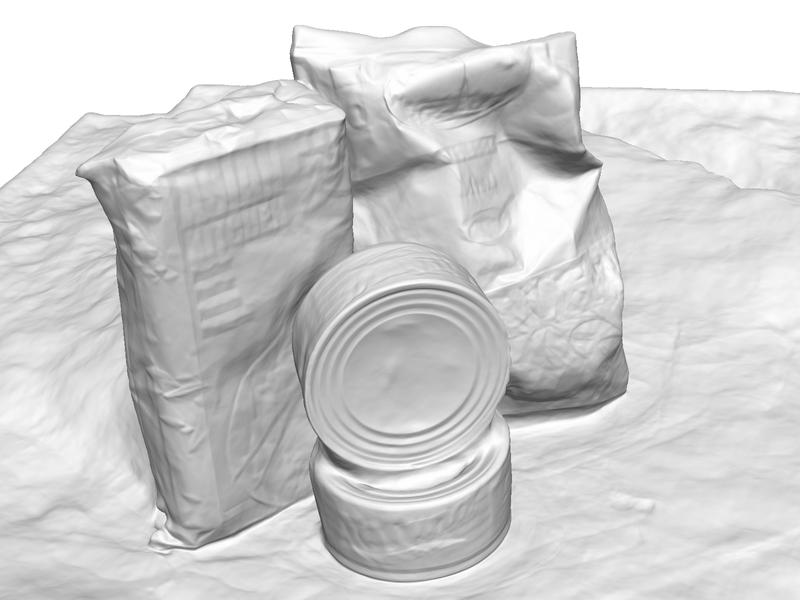}&
    \includegraphics[width=\mywidth]{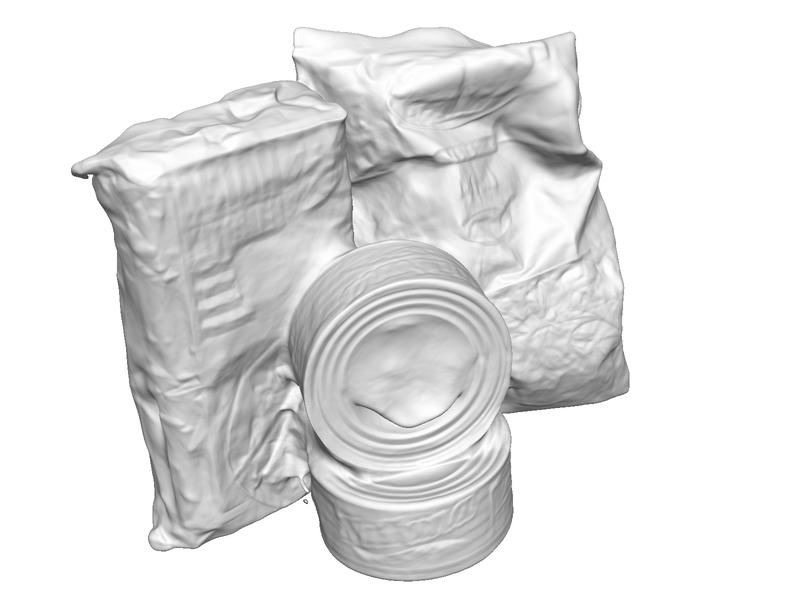}\\
    \includegraphics[width=\mywidth]{gfx/DTUcomp/GT_scan114.jpg}&
    \includegraphics[width=\mywidth]{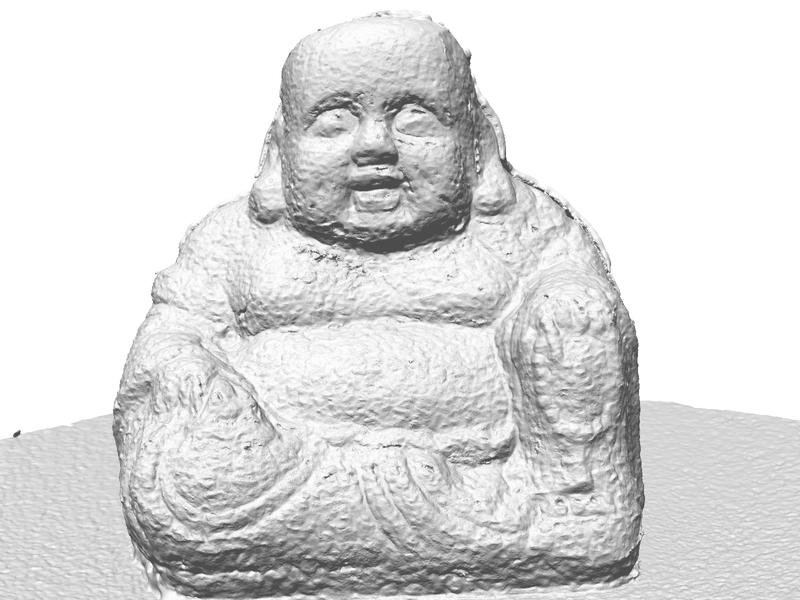}&
    \includegraphics[width=\mywidth]{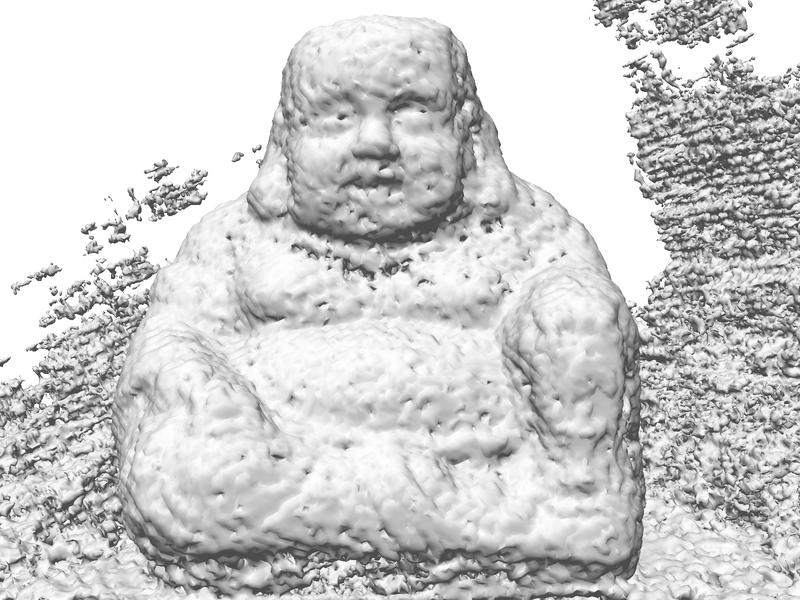}&
    \includegraphics[width=\mywidth]{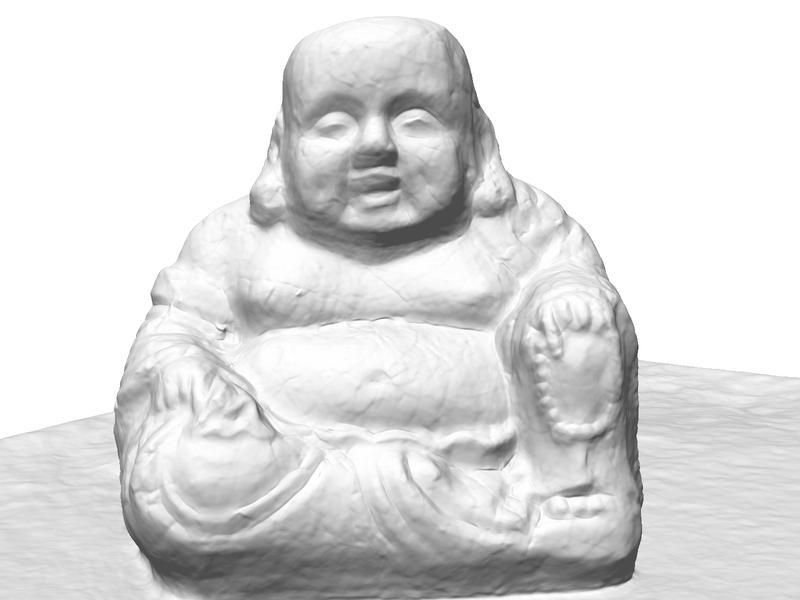}&
    \includegraphics[width=\mywidth]{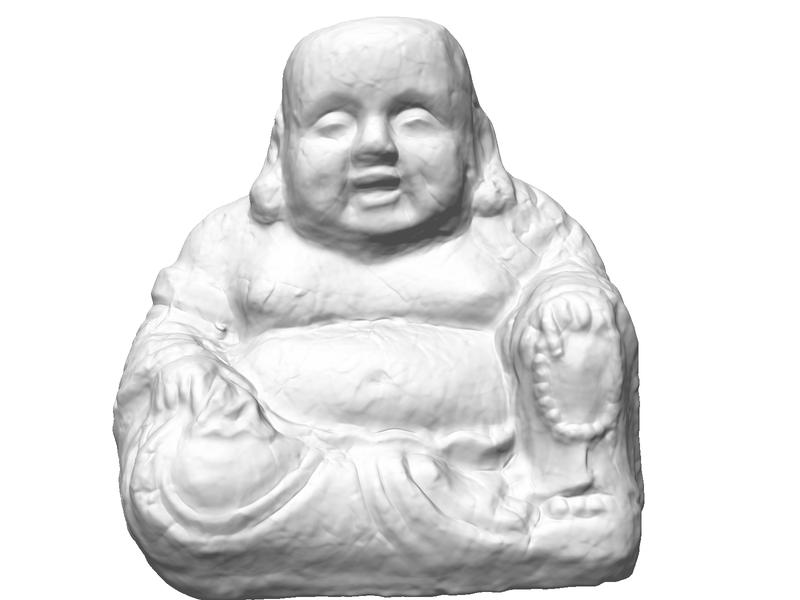}\\
    \includegraphics[width=\mywidth]{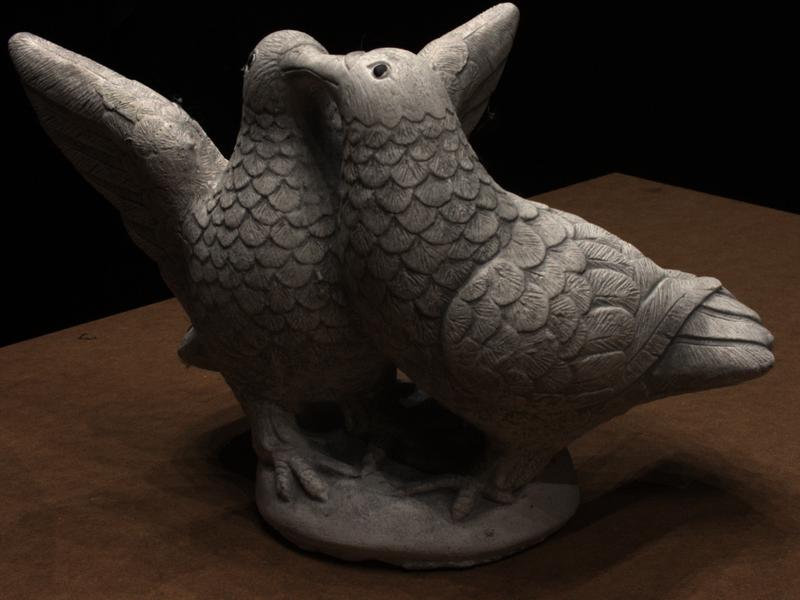}&
    \includegraphics[width=\mywidth]{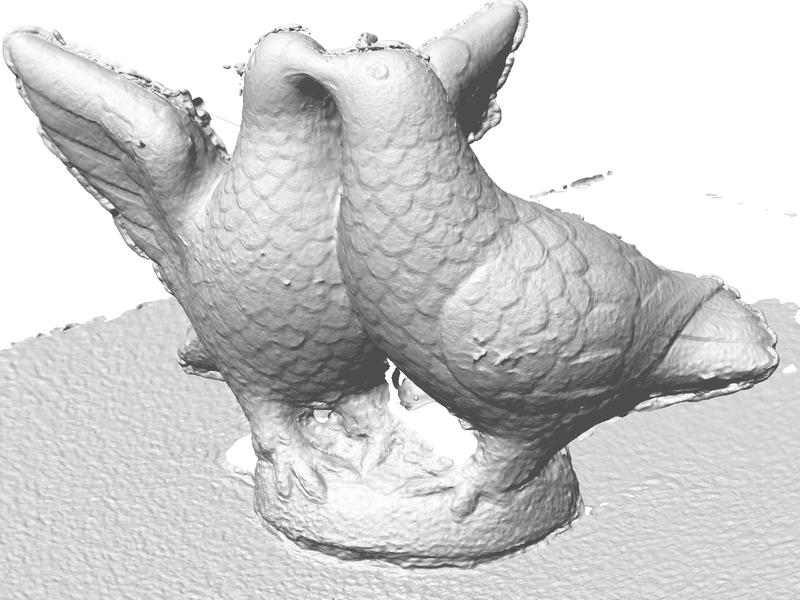}&
    \includegraphics[width=\mywidth]{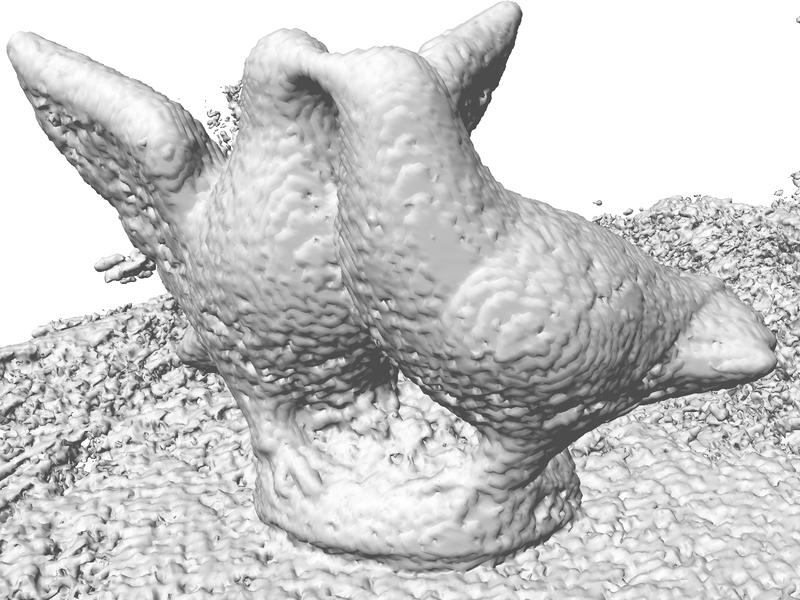}&
    \includegraphics[width=\mywidth]{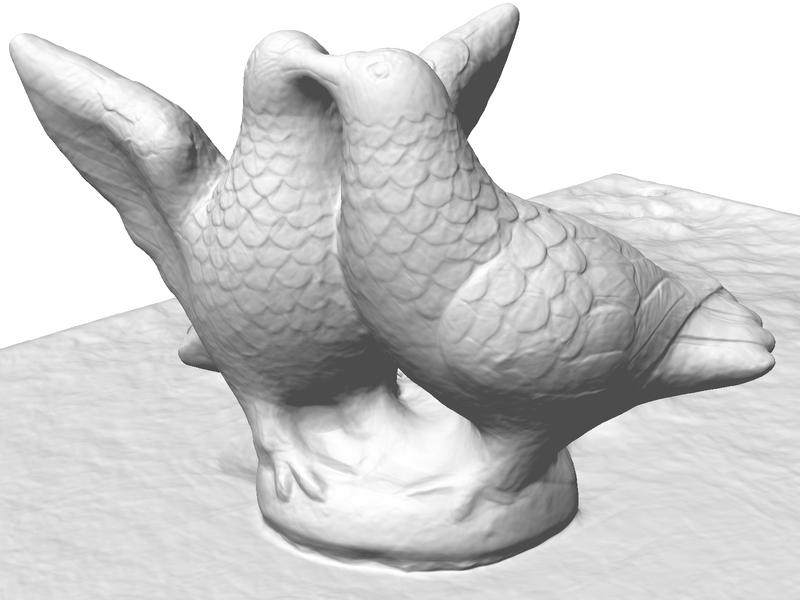}&
    \includegraphics[width=\mywidth]{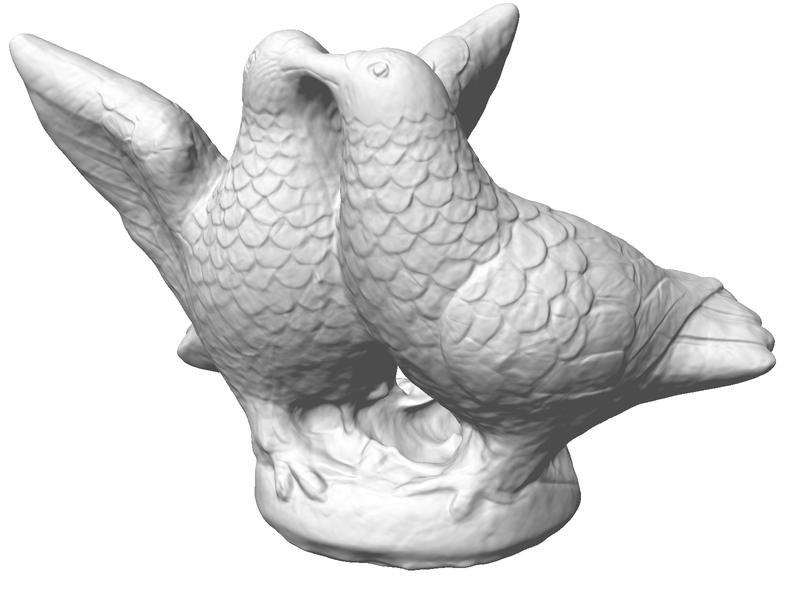}\\
    \includegraphics[width=\mywidth]{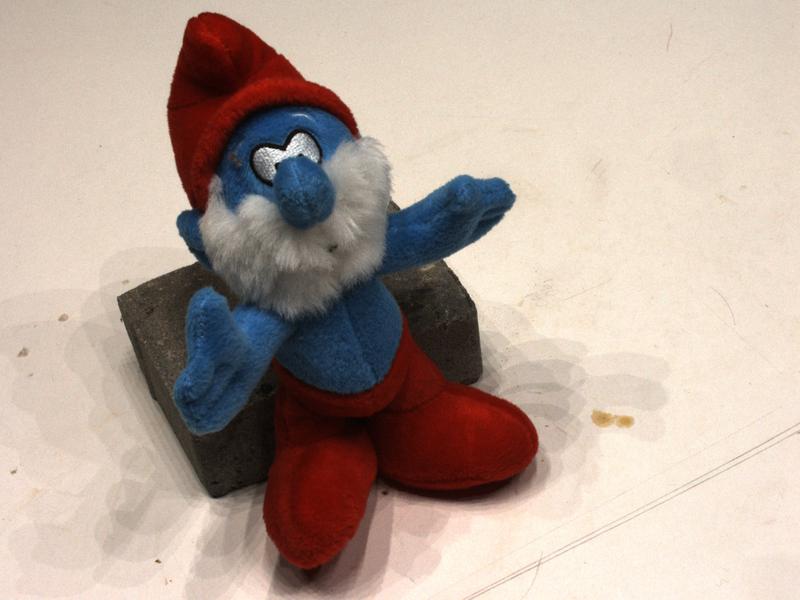}&
    \includegraphics[width=\mywidth]{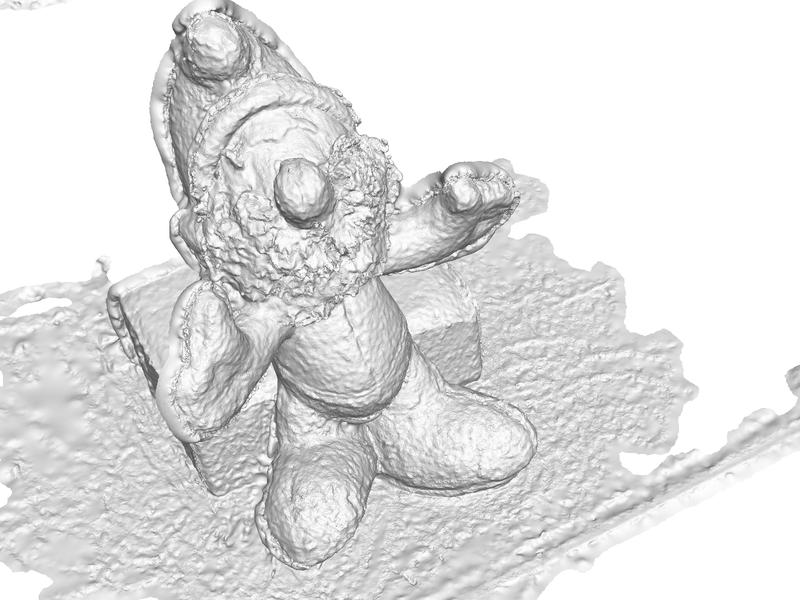}&
    \includegraphics[width=\mywidth]{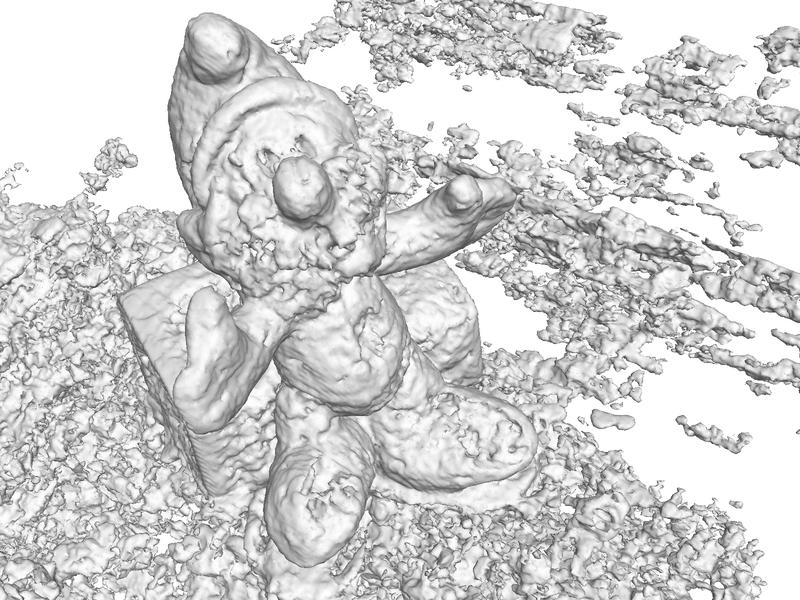}&
    \includegraphics[width=\mywidth]{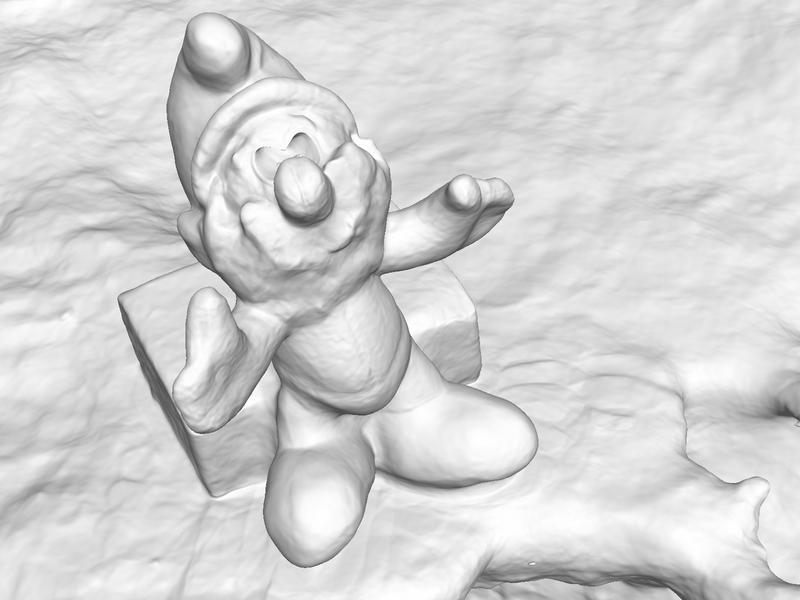}&
    \includegraphics[width=\mywidth]{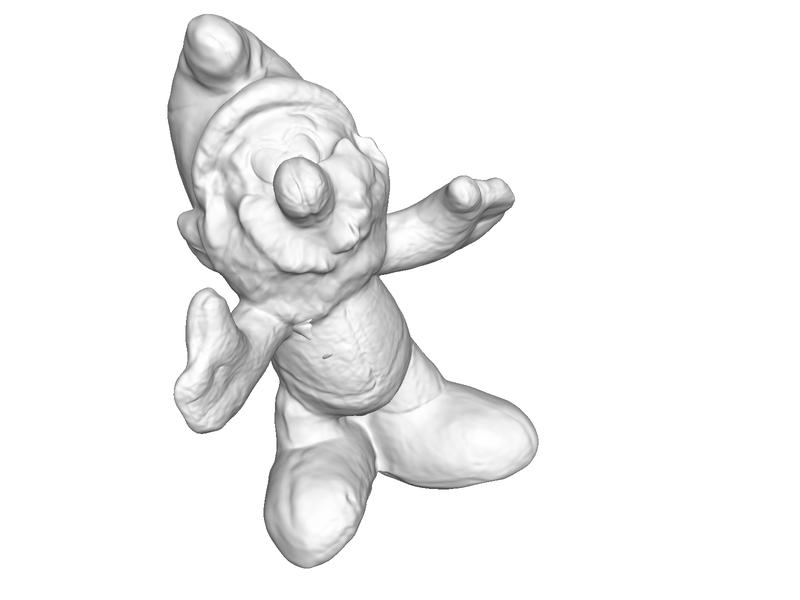}\\
         GT view&
         COLMAP \cite{Schoenberger2016ECCV}&
         NeRF \cite{Mildenhall2020ECCV} &
         \textbf{Ours} & IDR \cite{Yariv2020ARXIV}
         \\
    &$(\zeta=7)$&&&\\
\end{tabular}

%% file: tab/fig_3drec_indoor.tex
\begin{tabular}{@{}c@{}c@{}c@{}c@{}c}
    \includegraphics[width=0.25\linewidth]{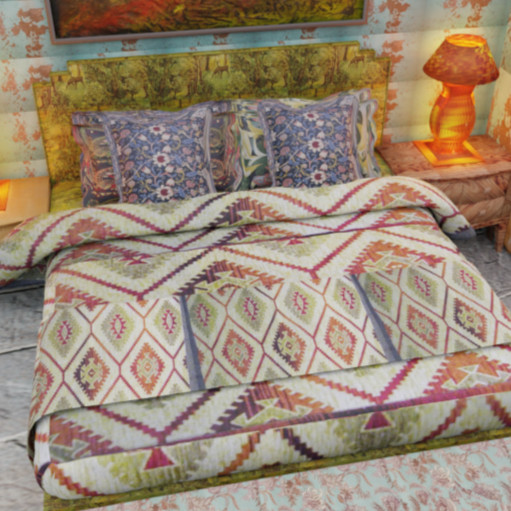}&
    \includegraphics[width=0.25\linewidth]{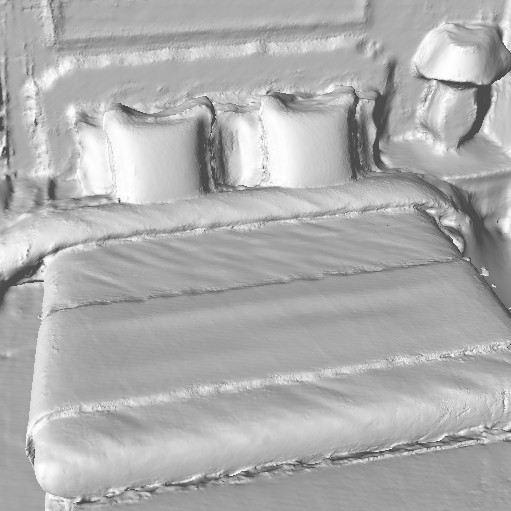}&
    \includegraphics[width=0.25\linewidth]{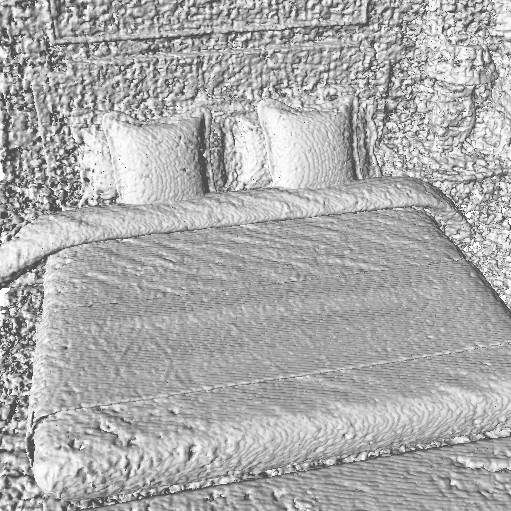}&
    \includegraphics[width=0.25\linewidth]{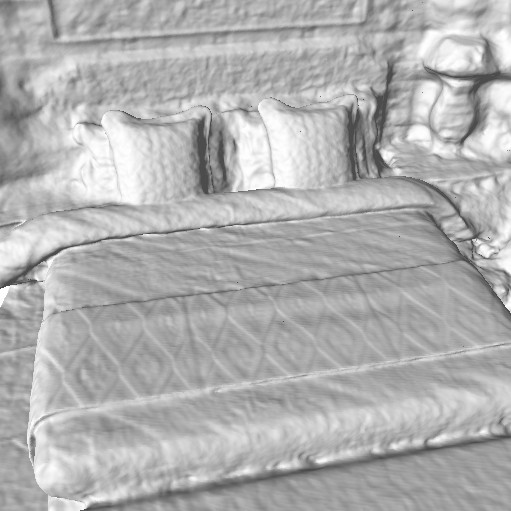}\\
    \includegraphics[width=0.25\linewidth]{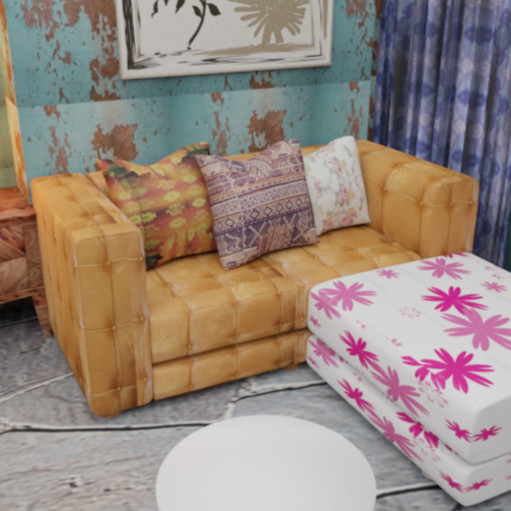}&
    \includegraphics[width=0.25\linewidth]{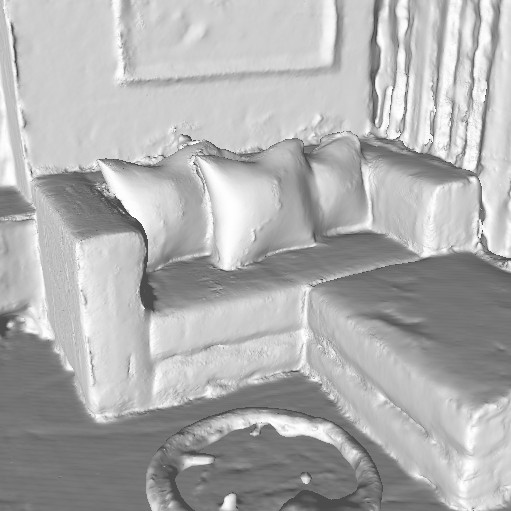}&
    \includegraphics[width=0.25\linewidth]{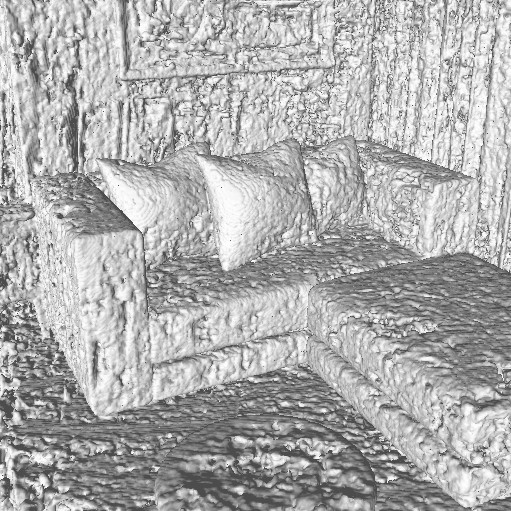}&
    \includegraphics[width=0.25\linewidth]{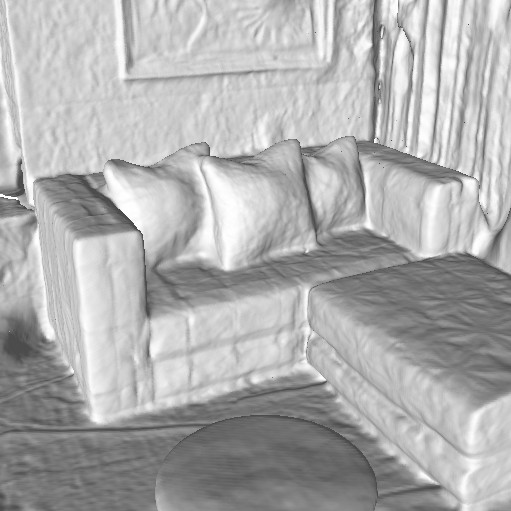}\\
    \includegraphics[width=0.25\linewidth]{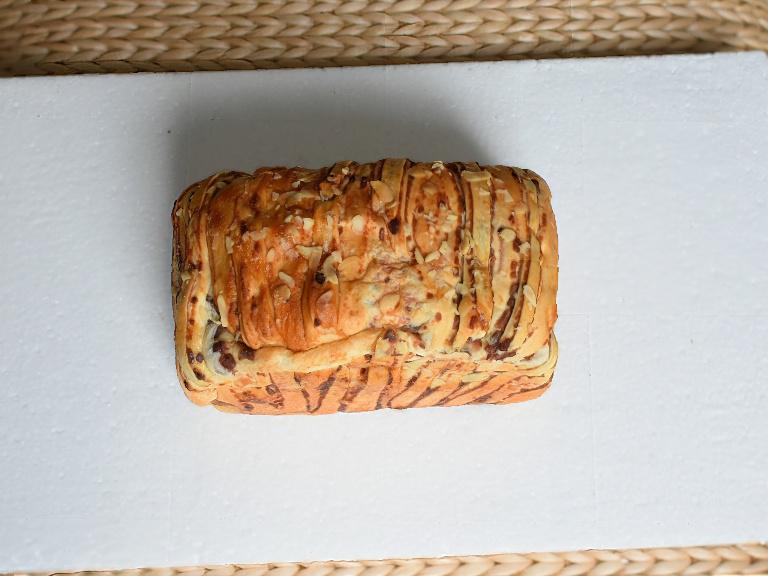}&
    \includegraphics[width=0.25\linewidth]{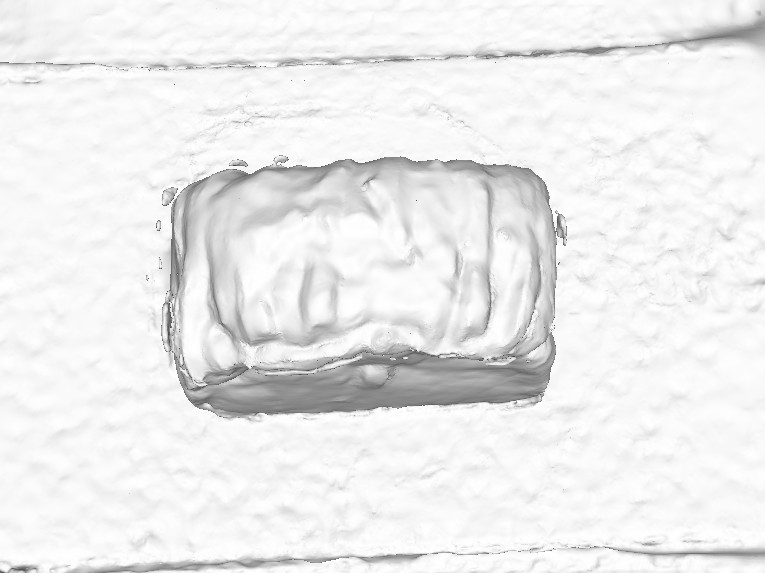}&
    \includegraphics[width=0.25\linewidth]{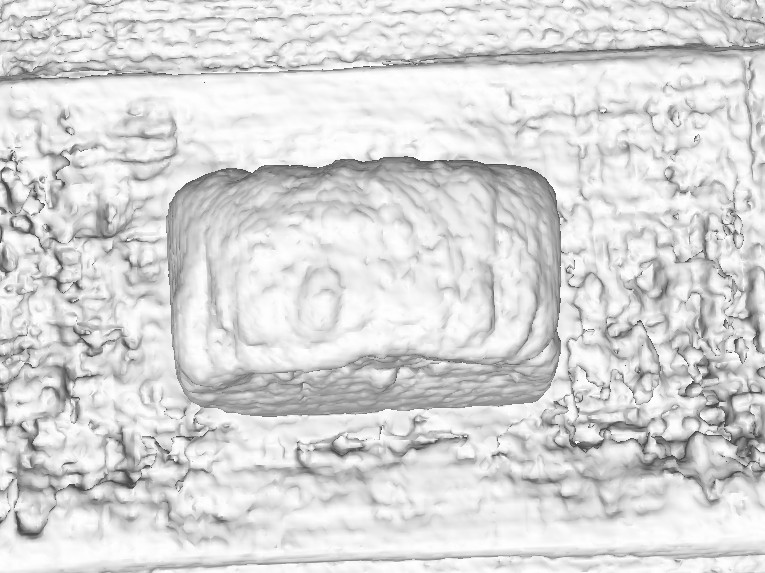}&
    \includegraphics[width=0.25\linewidth]{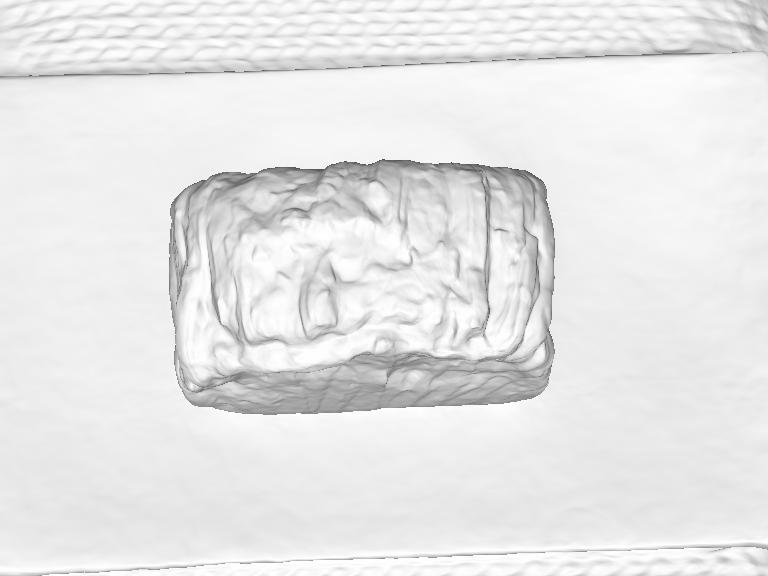}\\
    \includegraphics[width=0.25\linewidth]{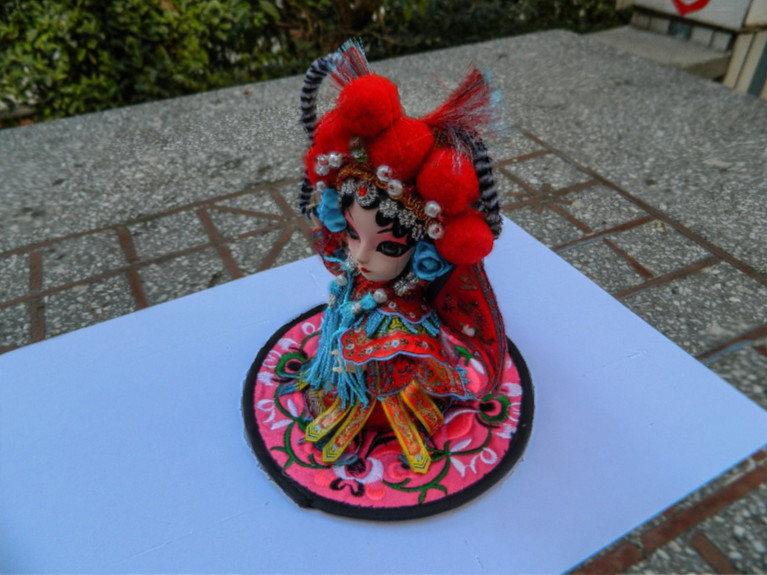}&
    \includegraphics[width=0.25\linewidth]{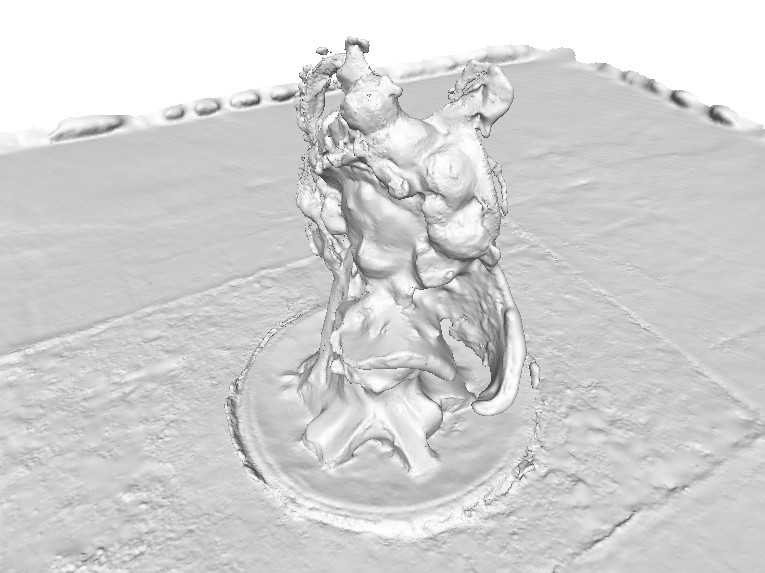}&
    \includegraphics[width=0.25\linewidth]{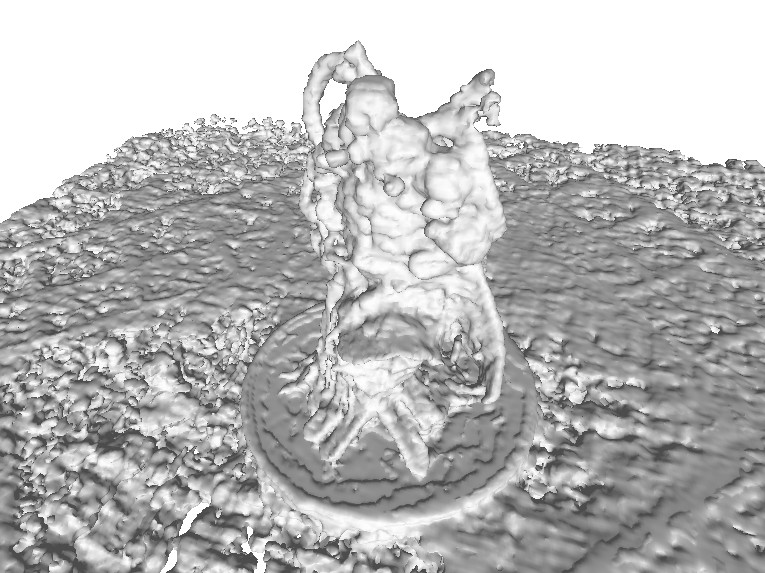}&
    \includegraphics[width=0.25\linewidth]{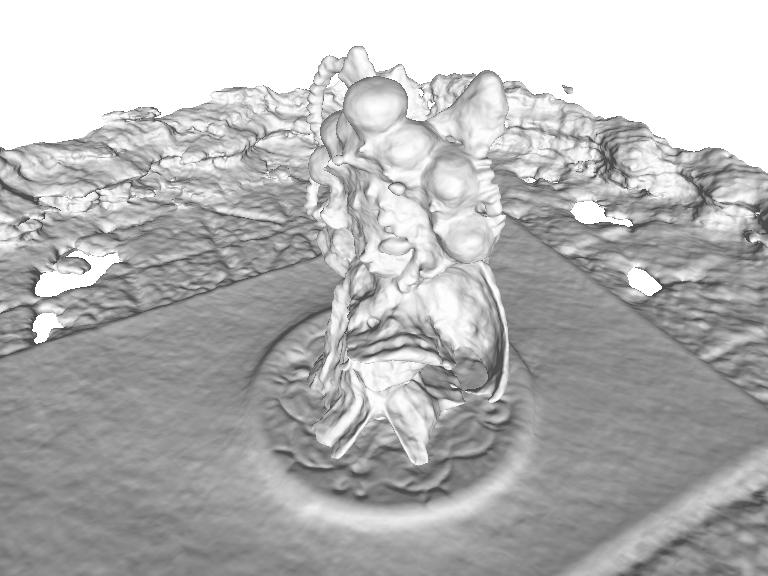}\\
         GT view&
         COLMAP \cite{Schoenberger2016ECCV}&
         NeRF \cite{Mildenhall2020ECCV} &
         Ours
         \\
\end{tabular}

%% file: sec_conclusion.tex
\section{Discussion and Conclusion}

This work presents UNISURF, a unified formulation of implicit surfaces and radiance fields for capturing high-quality implicit surface geometry from multi-view images without input masks.
We believe that neural implicit surfaces and advanced differentiable rendering procedures play a key role in future 3D reconstruction methods. 
Our unified formulation shows a path towards optimizing implicit surfaces in a more general setting than possible before.

\boldparagraph{Limitations}
By design, our model is limited to represent solid, non-transparent surfaces. Overexposed and textureless regions are also limiting factors that lead to inaccuraries and non-smooth surfaces. Furthermore, the reconstructions are less accurate at rarely visible regions in the images. Limitations are discussed in more detail in the supplementary.

In future work, for resolving ambiguities from rarely visible and texture-less regions, a prior is necessary for reconstruction.
While we incorporate an explicit smoothness prior during optimization, learning a probabilistic neural surface model which captures regularities and uncertainty across objects would help to resolve ambiguities, leading to more accurate reconstructions.
\vspace{-0.2cm}
\paragraph{Acknowledgement}
\small{
This work was supported by an NVIDIA research gift, the ERC Starting Grant LEGO-3D (850533) and the DFG EXC number 2064/1 - project number 390727645. Songyou Peng is supported by the Max Planck ETH Center for Learning Systems.}